\documentclass[10pt,twocolumn,letterpaper]{article}

\usepackage[pagenumbers]{cvpr} 
\usepackage{bm}
\usepackage{amsmath}
\usepackage{amssymb}
\usepackage{multirow} 

\usepackage{tabularx}
\usepackage{booktabs,multirow,graphicx,amssymb,xcolor}
\usepackage{array}
\usepackage{graphicx}
\usepackage{float}

\definecolor{copperref}{RGB}{0,128,184}

\usepackage{microtype}

\definecolor{cvprblue}{rgb}{0.21,0.49,0.74}
\usepackage[pagebackref,breaklinks,colorlinks,allcolors=cvprblue]{hyperref}
\usepackage[misc]{ifsym} 

\def\confName{CVPR}
\def\confYear{2027}

\title{Dynamic Forcing: Learning Action-Grounded Sparse Dynamic Representations with Dense Future Guidance for World Action Models}

\author{
Ziqiang Li\textsuperscript{1,4*}\quad \
Shisheng Chen\textsuperscript{1,4*}\quad \
Yunnan Wang\textsuperscript{1,4}\quad \
Wenyao Zhang\textsuperscript{1,4}\quad \
Zili Lin\textsuperscript{1,4}\quad \\
Yuyang Zhang\textsuperscript{1,4}\quad \
Xudong Yang\textsuperscript{1,4}\quad \
Fangzhuo Jin\textsuperscript{2}\quad \
Chenyu Zhou\textsuperscript{3,5}\quad \
Wenjun Zeng\textsuperscript{4}\quad \
Xin Jin\textsuperscript{4,5 \text{\Letter}} \\
\textsuperscript{1} Shanghai Jiao Tong University \quad\
\textsuperscript{2} Columbia University \quad\
\textsuperscript{3} Nanjing University \quad \\
\textsuperscript{4} Eastern Institute of Technology, Ningbo\quad \
\textsuperscript{5} Beijing Zhongguancun Academy
}

\usepackage[safe]{pbalance}

\begin{document}
\twocolumn[{
\renewcommand\twocolumn[1][]{#1}%
\maketitle
\vspace{-20pt}
\begin{center}
\vspace{-0.6cm}
\includegraphics[width=0.99\textwidth]{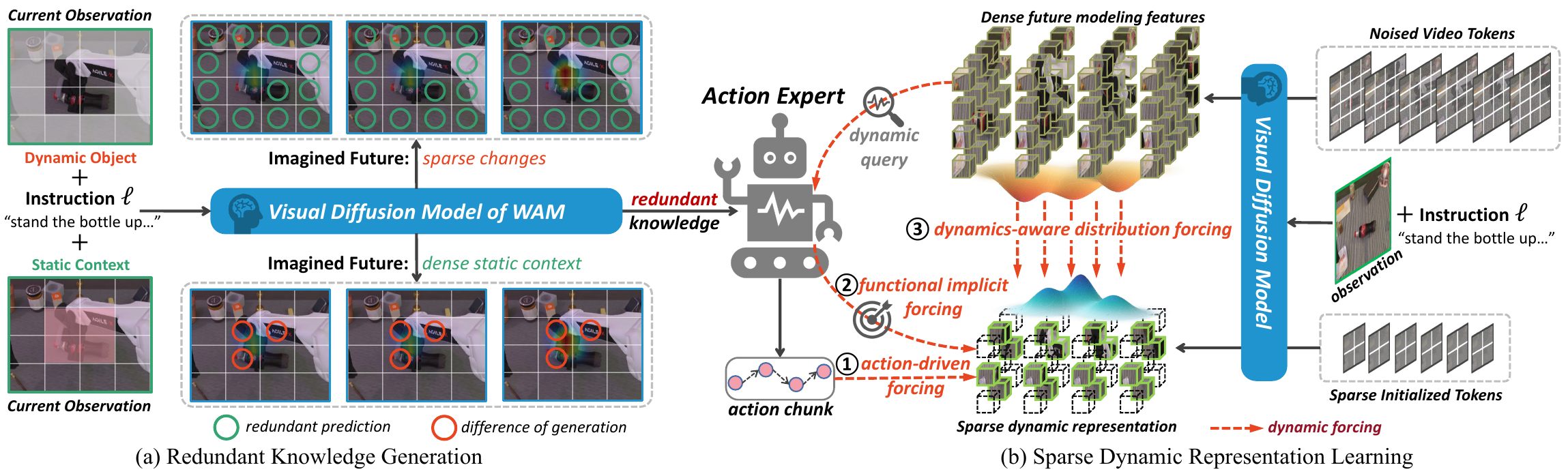}
\end{center}
\vspace{-0.6cm}
\captionof{figure}{\textbf{Sparse dynamic representation learning via Dynamic Forcing.} \textbf{(a)} Dense future modeling repeatedly encodes static context already present in the current observation, whereas action-expert attention concentrates on localized interaction regions. \textbf{(b)} Within CompactWAM's video-supervised shared backbone, Dynamic Forcing shapes compact dynamic representations through direct action-driven forcing (1) and privileged dense future guidance, comprising action functional guidance (2) and dynamics-aware distribution matching (3).}
\label{fig::teaser}
\vspace{0.15cm}
}]

\def\thefootnote{*}\footnotetext{Equal Contribution. \ \textsuperscript{\text{\Letter}} Corresponding author.}
\def\thefootnote{\arabic{footnote}} %

\maketitle
\begin{abstract}
World action models (WAMs) support robot control through visual prediction, but dense future modeling repeatedly encodes static context already available in the current observation, while action-relevant changes are often spatially sparse. We study sparse dynamic representation learning, retaining the current observation as a contextual anchor and using a small set of compact dynamic tokens to represent anticipated changes. We develop CompactWAM, a sparse-dense joint predictive framework that enables sparse representation learning within a video-supervised shared backbone, without requiring a reconstruction objective for the sparse pathway. Since a limited token budget alone does not specify which information to retain, we introduce Dynamic Forcing, which pairs direct action supervision with privileged dense future guidance to promote action relevance and dynamic fidelity. While action prediction gradients directly drive compact tokens to capture control-relevant cues, dense future guidance further shapes them via action functional guidance that matches sparse and dense contributions to the action expert, and dynamics-aware distribution matching, which aligns temporal relational changes relative to a static reference. At inference, a single visual forward pass encodes current observations and compact dynamic tokens for action generation, eliminating dense future modeling. CompactWAM achieves 98.4\% success on LIBERO and 77.9\% on LIBERO-Plus, retaining respective success rates of 98.3\% and 76.6\% with only 9.2\% of the dense future token budget. Evaluations on RoboTwin~2.0 and real-world manipulation further demonstrate the effectiveness of the learned sparse predictive representations.
\end{abstract}    
\section{Introduction}
\label{sec:intro}
Embodied intelligence requires models to understand their surroundings and anticipate the consequences of interaction. Vision-language-action (VLA) models advance this goal by transferring pretrained visual and linguistic knowledge to robot control, mapping observations and instructions to actions~\cite{rt2,openvla,pi0}. World action models (WAMs) incorporate visual dynamics into this process through generated visual plans and predictive features ~\cite{unipi,vpp}, as well as joint video-action modeling~\cite{uwm,uva}. These approaches provide policies with information about anticipated physical outcomes. However, modeling dense visual futures at deployment requires processing large sets of spatiotemporal tokens, introducing substantial inference overhead~\cite{fastwam}.

Recent work explores how to retain the benefits of predictive modeling at lower inference cost. Fast-WAM~\cite{fastwam} retains video co-training but removes test-time future prediction to improve inference efficiency. Subsequent evaluations show that removing future conditioning can weaken robustness under distribution shifts~\cite{fasterwam}. This raises a focused question: \emph{what should a WAM imagine beyond the current observation to support robot control without dense future prediction?} To investigate this, we revisit the content of dense future representations and how the action expert attends to them, as illustrated in Fig.~1(a). Dense futures repeatedly represent persistent appearance and scene context already available in the current observation, while the visualized action attention concentrates on localized regions associated with object motion and interaction. These observations motivate a complementary allocation of visual information: retain the current observation as a contextual anchor and devote sparse predictive representations to anticipated changes relevant to action generation.

We therefore investigate \emph{sparse dynamic representation learning} for WAMs: learning a limited set of compact dynamic tokens to represent anticipated changes while retaining the current observation as a contextual anchor. We formulate two complementary learning goals. \emph{\textbf{Dynamic fidelity}} requires these compact representations to capture temporal relational changes associated with future evolution. \emph{\textbf{Action relevance}} requires them to provide useful predictive context for action generation. A limited token budget alone, however, does not specify which information should be retained. The central challenge is thus to \emph{guide sparse representations to capture anticipated dynamics that support action generation}.

To this end, we develop \textbf{CompactWAM}, a sparse-dense joint predictive framework that enables sparse dynamic representation learning within a shared visual backbone. The dense pathway receives video and action supervision, providing a common predictive learning foundation, while the sparse pathway learns compact predictive representations without a video reconstruction objective of its own. Within this framework, we introduce \textbf{Dynamic Forcing}, a learning strategy that shapes sparse predictive representations through two complementary sources of supervision: \emph{direct action signals} and \emph{dense future modeling}. These sources jointly guide sparse representation learning toward action relevance and dynamic fidelity under a limited dynamic-token budget. First, \emph{direct action supervision} drives sparse representations to provide useful context for action generation. Action prediction gradients propagate through the sparse action expert into the sparse representation pathway, encouraging compact dynamic tokens to capture control-relevant information within the shared, video-supervised backbone. While establishing an action-driven baseline, action supervision alone does not explicitly specify which spatiotemporal structure to retain. Second, \emph{dense future supervision} provides privileged predictive knowledge to further regularize the sparse representations. As shown in Fig.~1(b), we transfer this knowledge through two complementary objectives. \emph{Action functional guidance} matches the isolated contributions of sparse dynamic and dense future representations in the attention readout space of the action expert, encouraging sparse context to retain the influence of future evolution on action generation. \emph{Dynamics-aware distribution matching} aligns distributions of spatiotemporal relational changes between sparse and dense representations, using a static relational reference to emphasize temporal evolution. Together with action supervision, these objectives guide sparse tokens to capture control-relevant dynamics.

In CompactWAM, we initialize compact dynamic tokens from pooled regions of the current observation and assign sparse spatiotemporal RoPE using region centroids and future-time coordinates, constructing a compact spatiotemporal representation space. During training, compact tokens learn through direct action supervision and guidance from the jointly trained dense pathway. At inference, only the current observation anchor and compact dynamic tokens are encoded in a single visual forward pass to provide predictive context for action generation. CompactWAM achieves success rates of 98.4\% on LIBERO and 77.9\% on LIBERO-Plus. Reducing the predictive context to just 9.2\% of the dense future token budget preserves strong performance, reaching 98.3\% and 76.6\%, respectively. These results support compact dynamic representations as effective predictive context for robust control.

Our contributions are summarized as follows:
\begin{itemize}
    \item We study sparse dynamic representation learning for WAMs by developing CompactWAM, enabling joint sparse-dense training and compact predictive conditioning alongside current observations at inference.
    \item We introduce Dynamic Forcing, combining direct action supervision with dense future knowledge transferred through action functional guidance and dynamics-aware distribution matching.
    \item Experiments on LIBERO~\cite{libero}, LIBERO-Plus~\cite{libero_plus}, RoboTwin~2.0~\cite{robotwin2}, and real-world manipulation demonstrate strong control performance and OOD robustness without dense future generation, validating sparse dynamic representations for robot control.
\end{itemize}

\section{Related Work}
\label{sec:related_work}

\begin{figure*}[!t]
\begin{center}
\includegraphics[width=1.\textwidth]{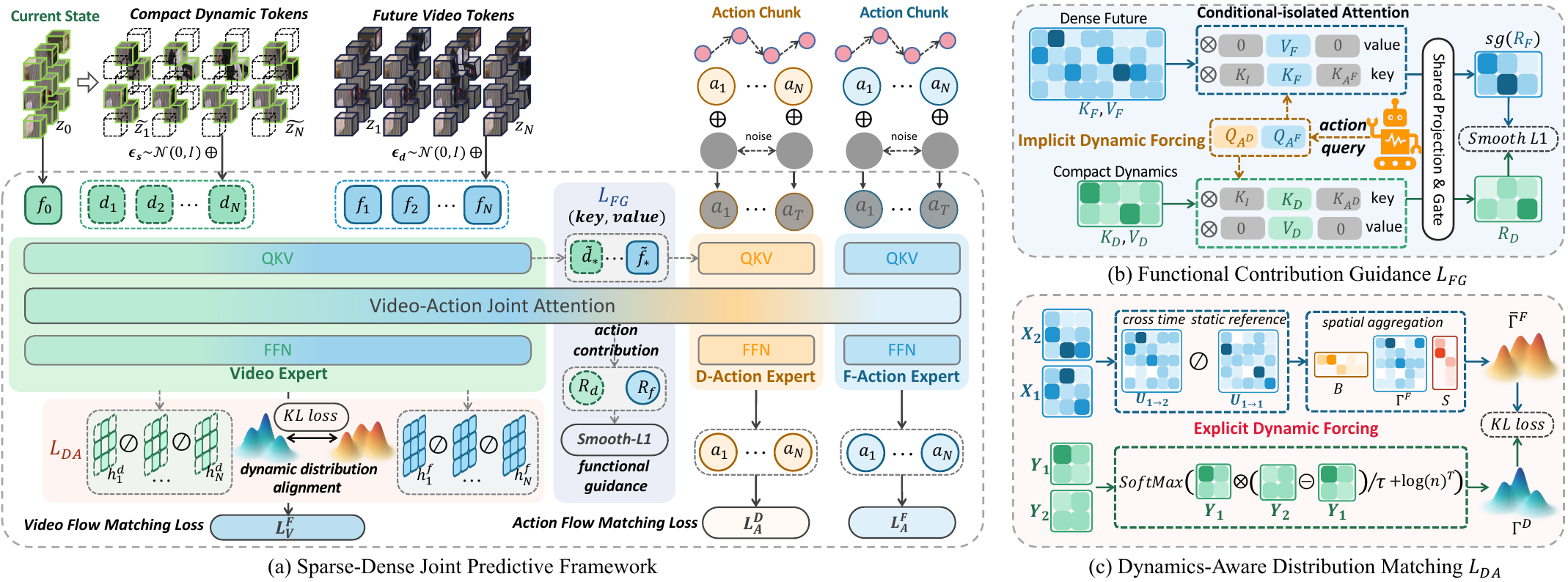}
\end{center}
\vspace{-0.62cm}
\caption{\textbf{Overview of Our Method.} \textbf{(a)} CompactWAM jointly learns dense futures and sparse dynamic representations within a video-supervised shared backbone, where direct action supervision ($\mathcal L_A^D$) provides action-driven forcing for the sparse pathway. Privileged dense future guidance complements this via \textbf{(b)} action functional guidance ($\mathcal L_{\mathrm{FG}}$), matching isolated predictive-context contributions to the action expert, and \textbf{(c)} dynamics-aware distribution matching ($\mathcal L_{\mathrm{DA}}$), aligning temporal relational changes.}
\vspace{-0.5cm}
\label{fig::mainpipeline}
\end{figure*}

\subsection{Vision-Language-Action Models}
Vision-language-action (VLA) models transfer pretrained visual and linguistic knowledge to robot control~\cite{rt2,openvla}. Their development is supported by increasingly diverse robot datasets~\cite{oxe,droid,agibot_world} and expressive generative action models~\cite{pi0,rdt,gr00t}. Recent studies improve open-world generalization and cross-embodiment transfer through heterogeneous training and adaptable policy representations~\cite{pi05,xvla,univla}. Complementary efforts explore structured action representations and unified reasoning--action generation~\cite{abotm0,fast_tokenizer,g05}. Efficient fine-tuning and asynchronous execution further improve the practicality of VLA deployment~\cite{openvla_oft,xiaomi_robotics0}. Our work investigates a complementary direction: learning compact predictive representations of visual dynamics to support action generation.

\subsection{World Action Models}
World action models (WAMs) incorporate visual dynamics into policy learning, building on video-based control through generated plans and predictive visual features~\cite{unipi,seer,vpp,vidar}. Another line integrates visual prediction and action generation within unified autoregressive or diffusion formulations~\cite{uwm,worldvla,uva,motus,cosmos_policy,lingbot_va}. Across these approaches, visual prediction provides both a learning signal for world representations and a possible source of context for action generation. Recent work examines how these benefits can be retained with lower deployment cost. Fast-WAM~\cite{fastwam} retains video co-training while removing test-time future prediction, whereas Faster-WAM~\cite{fasterwam} preserves future conditioning through a single video-expert pass and selective video--action interaction. Other approaches reduce computation through compact video backbones, efficient action decoding, or token-sparse future latents~\cite{light_wam,efficient_wam}. AHA-WAM~\cite{ahawam} further amortizes world-model computation through asynchronous planner--executor execution and observation-guided routing of cached context. 

We focus on \emph{sparse dynamic representation learning}: learning compact representations of anticipated changes under a limited predictive-token budget, with the current observation as a contextual anchor. Within CompactWAM's video-supervised shared backbone, \emph{Dynamic Forcing} combines direct action supervision with privileged dense future guidance. The latter transfers temporal relational changes and the contributions of dense predictive context to action generation. At deployment, compact dynamic tokens provide predictive context for action generation without dense future prediction.

\section{Method}
\label{sec:method}

\noindent\textbf{Paradigm formulation.} Distinct from WAMs that reconstruct dense visual futures for robot control, we learn sparse dynamic representations within a sparse--dense joint predictive framework. During training, given the current observation $\mathbf I$ and language instruction $\ell$, we jointly model dense future representations $\mathbf F$, sparse dynamic representations $\mathbf D$, and their respective action chunks, $\mathbf{A}^F$ and $\mathbf{A}^D$:
{\setlength{\abovedisplayskip}{2pt}\setlength{\belowdisplayskip}{2pt}
\begin{equation}
\begin{aligned}
    & p_\theta(\mathbf{A}^F, \mathbf{A}^D, \mathbf F, \mathbf D \mid \mathbf I, \ell) \\
     = & \underbrace{
        p_\theta(\mathbf{A}^F, \mathbf{A}^D \mid \mathbf F, \mathbf D, \mathbf I, \ell)
    }_{\text{action guidance}}
    \underbrace{
        p_\theta(\mathbf F, \mathbf D \mid \mathbf I, \ell)
    }_{\text{joint modeling}},
\end{aligned}
\label{eq:sw_train_policy}
\end{equation}   
}%
The action term comprises two independent action experts conditioned on $[\mathbf I,\mathbf F,\ell]$ and $[\mathbf I,\mathbf D,\ell]$, respectively. The predictive pathways share a visual backbone trained with dense video prediction, while the sparse pathway learns compact predictive representations through action supervision without its own video reconstruction objective. Within this architecture, \emph{Dynamic Forcing} combines two complementary sources of supervision: direct action signals and privileged guidance from dense future modeling. After joint training, inference retains only the sparse pathway:
{\setlength{\abovedisplayskip}{4pt}\setlength{\belowdisplayskip}{4pt}
\begin{equation}
\begin{aligned}
     p_\theta(\mathbf A^D, \mathbf D \mid \mathbf I, \ell)
     = \underbrace{p_\theta(\mathbf A^D \mid\mathbf D, \mathbf I, \ell)}_{\text{action guidance}}
        \underbrace{p_\theta(\mathbf D \mid \mathbf I, \ell)}_{\text{sparse evolution}}.
\end{aligned}
\label{eq:sw_eval_policy}
\end{equation}  
}%
The current observation provides rich scene context, while compact dynamic tokens supply learned predictive context without dense future generation.

\noindent\textbf{Overview.}
As shown in Figure~\ref{fig::mainpipeline}, CompactWAM jointly learns sparse dynamic representations and dense futures within a video-supervised shared backbone, with separate action experts. Compact tokens are initialized from current observations with sparse 3D-RoPE coordinates to preserve spatiotemporal organization. Dynamic Forcing combines \emph{direct action supervision}, which drives control-relevant sparse representation learning through action prediction gradients, and \emph{privileged dense future guidance}. The latter comprises action functional guidance (FG), matching sparse and dense context contributions to the action expert, and dynamics-aware distribution matching (DA), aligning temporal relational changes. At inference, current-observation features and sparse dynamic representations guide action generation without dense future modeling.

\subsection{Sparse-Dense Joint Predictive Framework}
\label{sec:compactwam_framework}
We develop a sparse-dense joint predictive framework that supports action-driven learning of sparse dynamic representations. As depicted in Figure~\ref{fig::mainpipeline}(a), the two predictive pathways share a visual backbone and each has its own action expert. Dense video supervision provides a common predictive learning foundation, while the sparse action objective trains compact representations for control. This design enables sparse representation learning without a video reconstruction objective for the sparse pathway and supports additional supervision from dense future predictions.

\noindent\textbf{Compact dynamic token initialization.} We initialize compact dynamic tokens from the current observation. Let $N$ be the number of spatial tokens per latent slice of $\mathbf I$. We partition the observed token grid into $R$ non-overlapping local regions, where $R\ll N$. The assignment matrix $\bm S\in\{0,1\}^{N\times R}$ maps each dense token to one region, with $\bm S_{ir}=\mathbf 1[i\in\mathcal R_r]$. The regional averaging matrix is
{\setlength{\abovedisplayskip}{4pt}\setlength{\belowdisplayskip}{4pt}
\begin{equation}
    \bm B=\operatorname{diag}(n)^{-1}\bm S^\top,
    \label{eq:sw_fixed_maps}
\end{equation}
}%
where $n_r=\sum_i\bm S_{ir}$ and $\bm B\in\mathbb R^{R\times N}$ averages tokens within each region. Using the same partition and slot ordering for future latent time $f$, we initialize dynamic tokens as
{\setlength{\abovedisplayskip}{4pt}\setlength{\belowdisplayskip}{4pt}
\begin{equation}
    \mathbf D_f
    =(1-s)\cdot\bm B\operatorname{sg}(\mathbf I)
    +s\cdot\epsilon_D,
    \label{eq:sw_planning_init}
\end{equation}
}%
where $\epsilon_D$ is Gaussian noise, $\operatorname{sg}$ denotes stop-gradient, and $s\in[0,1]$ is the normalized noise timestep shared with the dense pathway. Regional pooling provides a spatially grounded initialization, while noise interpolation connects compact dynamic tokens to the video backbone's denoising formulation. For each future latent slice, we assign each slot the spatial centroid of its source region and the corresponding future-time coordinate. These continuous coordinates, computed as $\bm{\overline U}=\bm B\bm U$ from the dense spatiotemporal grid, define sparse 3D-RoPE and preserve the spatial organization and temporal identity of compact dynamic tokens.

\noindent\textbf{Attention masking.}
During training, the visual backbone processes $[\mathbf I,\mathbf F,\mathbf D]$. Current-observation tokens attend only to $\mathbf I$, dense future tokens attend to $[\mathbf I,\mathbf F]$, and compact dynamic tokens attend to $[\mathbf I,\mathbf D]$. This mask prevents sparse tokens from accessing dense future features, including indirectly through the current-observation tokens.

\noindent\textbf{Action-driven forcing.}
Dense video prediction trains the shared backbone to model visual dynamics, while action supervision shapes compact dynamic representations for control. The sparse action expert predicts demonstrated actions conditioned on $[\mathbf I,\mathbf D,\ell]$, supervised by $\mathcal L_A^D$ in Eq.~\eqref{eq:sw_flow_loss}. Its gradients propagate through the action expert and sparse visual pathway into the shared backbone, encouraging compact dynamic tokens to encode information useful for action generation. We term this supervision from demonstrated actions \emph{action-driven forcing}. Within the joint framework, it is complemented by privileged guidance from the dense pathway: FG transfers the functional contribution of predictive context to action generation, and DA aligns temporal relational changes. Together, direct action supervision and dense future guidance constitute Dynamic Forcing, shaping compact representations toward action relevance and dynamic fidelity.

\subsection{Forcing in Action Functional Guidance}
\label{sec:sparsewam_fa}
Beyond direct action supervision, dense future modeling provides a privileged functional target. The action expert's attention readout measures how predictive context contributes to action generation, providing a functional criterion for the information compact tokens should retain. Within our joint framework, we use the contribution of dense future context to the action expert as a supervision target for sparse dynamic representations. Matching their isolated attention contributions encourages sparse tokens to preserve the action-guiding function of dense futures, providing \emph{implicit dynamic forcing} through the action expert's functional readout. 

For a block $k\in\mathcal K_c$ and predictive context $Z\in\{D,F\}$, let $\bm Q_{A^Z}^k$ and $\bm K_{A^Z}^k$ denote the queries and keys of the corresponding action expert. Define the concatenated keys $\widetilde{\bm K}_Z^k=[\operatorname{sg}(\bm K_I^k);\bm K_Z^k;\operatorname{sg}(\bm K_{A^Z}^k)]$ and masked values $\widetilde{\bm V}_Z^k=[\bm 0;\bm V_Z^k;\bm 0]$. We isolate the contribution of predictive context as
{\setlength{\abovedisplayskip}{4pt}\setlength{\belowdisplayskip}{4pt}
\begin{equation}
    \bm G_Z^k
    =\operatorname{SoftMax}\!\left(
        \frac{
            \operatorname{sg}(\bm Q_{A^Z}^k)
            (\widetilde{\bm K}_Z^k)^\top
        }{\sqrt{d_k}}
    \right)\widetilde{\bm V}_Z^k.
\label{eq:sw_condition_contribution}
\end{equation}
}%
Here, $d_k$ is the attention-head dimension. Each pathway uses its own action query and attention denominator. We zero the current-observation and action values to isolate the predictive contribution. Both contributions are then mapped into the sparse action expert's residual space: $\bm R_Z^k=(\bm G_Z^k\operatorname{sg}(\bm W_O^k)^\top)\odot\operatorname{sg}(\bm\gamma_A^k)$, where $\bm W_O^k$ and $\bm\gamma_A^k$ are its output projection and timestep-dependent residual gate. The functional guidance loss is
{\setlength{\abovedisplayskip}{4pt}\setlength{\belowdisplayskip}{4pt}
\begin{equation}
    \mathcal L_{\mathrm{FG}}
    =\frac{1}{|\mathcal K_c|}
    \sum_{k\in\mathcal K_c}
    \operatorname{SmoothL1}\!\left(
        \bm R_D^k,\operatorname{sg}(\bm R_F^k)
    \right),
    \label{eq:sw_fa_loss}
\end{equation}
}%
where each block loss averages over valid action queries and feature dimensions. We detach the dense target, output projection, and residual gate, allowing the matching objective to guide sparse representation learning. Dense future modeling thus supplies predictive knowledge, while the action expert defines the functional form in which that knowledge complements direct action supervision.

\subsection{Dynamics-Aware Distribution Forcing}
\label{sec:sparsewam_da}
While FG transfers the functional contributions of dense predictive context, we complement it with explicit supervision of temporal relational changes. Cross-time affinities encode both persistent scene structure and temporal variation. To emphasize the latter, we construct a relation-gain distribution that characterizes dynamics in cross-time affinities relative to a within-time reference. Aligning sparse representations with this dynamics-aware distribution provides \emph{explicit dynamic forcing} from dense future modeling, encouraging them to preserve its temporal relational structure.

\noindent\textbf{Dynamic relations relative to a static reference.} We extract dense future and sparse dynamic hidden features $H^F$ and $H^D$ from selected visual blocks. At a selected block, let $X_t=\phi(H_t^F)$ and $Y_t=\phi(H_t^D)$, where $X_t\in\mathbb R^{N\times d_v}$ and $Y_t\in\mathbb R^{R\times d_v}$. Here, $d_v$ denotes the visual feature dimension. The mapping $\phi$ applies LayerNorm followed by $\ell_2$ normalization, and $t$ indexes future latent-time groups. Consider two groups, indexed by $1$ and $2$, with consistent spatial token ordering. With temperature $\tau$, define
{\setlength{\abovedisplayskip}{4pt}\setlength{\belowdisplayskip}{4pt}
\begin{equation}
\begin{aligned}
    \bm U_{1\rightarrow2}
    &=\operatorname{SoftMax}
    (\bm X_1\bm X_2^\top/\tau),\\
    \bm U_{1\rightarrow1}
    &=\operatorname{SoftMax}
    (\bm X_1\bm X_1^\top/\tau).
\end{aligned}
\label{eq:sw_relation_reference}
\end{equation}
}%
Here, $\bm U_{1\rightarrow2}$ captures cross-time affinities, while $\bm U_{1\rightarrow1}$ provides a within-time reference. To emphasize changes relative to this reference, we define the relation-gain distribution as $\bm\Gamma^F=\operatorname{Norm}(\bm U_{1\rightarrow2}\oslash\bm U_{1\rightarrow1})$, where $\oslash$ denotes element-wise division and $\operatorname{Norm}$ normalizes each row. Canceling the row-wise normalization constants gives
{\setlength{\abovedisplayskip}{4pt}\setlength{\belowdisplayskip}{4pt}
\begin{equation}
    \bm\Gamma^F
    =\operatorname{SoftMax}\!\left(
        \frac{\bm X_1(\bm X_2-\bm X_1)^\top}{\tau}
    \right).
    \label{eq:sw_teacher_gain}
\end{equation}
}%
The $(i,j)$ logit measures the change in affinity between source token $i$ and destination location $j$ relative to the within-time reference.

\noindent\textbf{Dynamics-aware distribution matching.} To construct a supervision target compatible with the sparse dynamic slots, we aggregate the dense relation-gain distribution using the averaging matrix in Eq.~\eqref{eq:sw_fixed_maps} as $\bm{\overline\Gamma}^F=\bm B\bm\Gamma^F\bm S$. Here, $\bm B$ averages source-token rows within each region, while $\bm S$ accumulates destination probability mass over each destination region. The sparse relation-gain distribution is
{\setlength{\abovedisplayskip}{4pt}\setlength{\belowdisplayskip}{4pt}
\begin{equation}
    \bm\Gamma^D_{rs}
    =
    \frac{
        n_s\exp\!\left(
            \bm Y_{1,r}^\top
            (\bm Y_{2,s}-\bm Y_{1,s})/\tau
        \right)
    }{
        \sum_{i=1}^{R}n_i\exp\!\left(
            \bm Y_{1,r}^\top
            (\bm Y_{2,i}-\bm Y_{1,i})/\tau
        \right)
    },
    \label{eq:sw_p_gain}
\end{equation}
}%
where $\bm Y_{t,r}$ is the feature of sparse slot $r$ at time $t$. The factor $n_s$ weights destination slots by their region sizes to match the regional probability mass in the aggregated dense target. The dynamics-aware distribution (DA) matching loss is
{\setlength{\abovedisplayskip}{4pt}\setlength{\belowdisplayskip}{4pt}
\begin{equation}
    \mathcal L_{\mathrm{DA}}
    =
    \frac{1}{|\mathcal K_d|}
    \sum_{l\in\mathcal K_d}
    D_{\mathrm{KL}}\!\left(
        \operatorname{sg}(\bm{\overline\Gamma}^{F,l})
        \parallel\bm\Gamma^{D,l}
    \right),
    \label{eq:sw_gain_loss}
\end{equation}
}%
where $\mathcal K_d$ denotes the selected visual blocks, and the KL divergence is computed row-wise and averaged over source regions. We detach the aggregated dense target and guide sparse dynamic representation learning through temporal relations. Together, FG and DA provide dense future guidance that complements action-driven forcing: FG transfers functional contributions to action generation, while DA transfers temporal relational changes. 

\subsection{Joint Training and Inference}
\label{sec:sparsewam_training}
For the video pathway and the two action experts, indexed by $m\in\{V,A^D,A^F\}$, the flow-matching objective is
{\setlength{\abovedisplayskip}{4pt}\setlength{\belowdisplayskip}{4pt}
\begin{equation}
\begin{aligned}
    \mathcal L_m=\mathbb E\!\left[
\bigl\|v_\theta^m(x_t^{m},\mathcal C_m)-(\epsilon_m-x_0^m)\bigr\|_2^2
      \right],
\end{aligned}
    \label{eq:sw_flow_loss}
\end{equation}
}%
where $\mathcal C_m$ contains the noise timestep and the permitted conditioning inputs. The sparse and dense action experts are conditioned on $[\mathbf I,\mathbf D,\ell]$ and $[\mathbf I,\mathbf F,\ell]$, respectively. The video objective supervises dense future latents conditioned on $\mathbf I$ and $\ell$. The full training objective is:
{\setlength{\abovedisplayskip}{4pt}\setlength{\belowdisplayskip}{4pt}
\begin{equation}
\begin{aligned}
    \mathcal L= \underbrace{\mathcal L_V+\mathcal L_A^F}_{\text{Predictive Learning}} +\underbrace{\mathcal L_A^D +\lambda\cdot \mathcal L_{\mathrm{FG}}+\beta\cdot \mathcal L_{\mathrm{DA}}}_{\text{Dynamic Forcing}}.
\end{aligned}
    \label{eq:sw_total_loss}
\end{equation}
}%
Here, $\mathcal L_V$ and $\mathcal L_A^F$ train the dense pathway, while Dynamic Forcing combines direct action supervision $\mathcal L_A^D$ with privileged dense guidance $\lambda\mathcal L_{\mathrm{FG}}+\beta\mathcal L_{\mathrm{DA}}$, weighted by $\lambda$ and $\beta$. At inference, a single visual forward pass encodes $[\mathbf I,\mathbf D]$ at the configured denoising timestep conditioned on $\ell$, caching keys and values across action denoising steps.

\section{Experiments}
\label{sec:experiment}
Our experiments address three questions: (i) whether CompactWAM can retain the control performance and robustness of dense future modeling using sparse dynamic representations, (ii) how direct action supervision and privileged dense future guidance contribute to sparse representation learning at a fixed token budget, and (iii) how the predictive token budget affects the trade-off between control performance and inference cost.

\begin{table*}[t]
  \caption{\textbf{Success rates (\%) on LIBERO and LIBERO-Plus.} Emb.~PT. denotes embodied pretraining prior to benchmark fine-tuning, and Future Ratio indicates the proportion of predictive visual tokens processed at inference relative to full dense future generation. Best results within each paradigm are shown in \textbf{bold}.}
  \vspace{-0.2cm}
  \label{tab:Libero}
  \centering
  \small
  \setlength{\tabcolsep}{3pt}
  \renewcommand{\arraystretch}{0.90}

  \resizebox{\textwidth}{!}{%
    \begin{tabular}{l|cc|ccccc|cccccccc}
      \toprule[0.9pt]

      \multirow{2}{3.0cm}{Method}
      & \multirow{2}{*}{\shortstack{\textbf{Emb.}\\\textbf{PT.}}}
      & \multirow{2}{*}{\shortstack{\textbf{Future}\\\textbf{Ratio}}}
      & \multicolumn{5}{c|}{LIBERO}
      & \multicolumn{8}{c}{LIBERO-Plus} \\

      \cmidrule(lr){4-8}
      \cmidrule(l){9-16}

      & & &
      Long & Goal & Object & Spatial & Avg.
      & Camera & Robot & Language & Light
      & Background & Noise & Layout & Avg. \\

      \midrule
      \multicolumn{16}{l}{\textit{VLA}} \\
      \midrule

      $\pi_{0.5}$~\cite{pi05}
      & $\checkmark$ & -
      & 92.4 & \textbf{98.0} & 98.2 & \textbf{98.8} & 96.9
      & \textbf{75.4} & 77.5 & \textbf{85.6} & \textbf{96.9}
      & 94.6 & \textbf{89.7}
      & \textbf{85.7} & \textbf{85.7} \\

      UniVLA~\cite{univla}
      & $\checkmark$ & -
      & 92.0 & 95.6 & 96.8 & 96.5 & 95.2
      & 1.8 & 46.2 & 69.6 & 69.0
      & 81.0 & 21.2 & 31.9 & 42.9 \\

      OpenVLA-OFT~\cite{openvla_oft}
      & $\checkmark$ & -
      & 94.5 & 97.9 & 98.4 & 97.6 & 97.1
      & 56.4 & 31.9 & 79.5 & 88.7
      & 93.3 & 75.8 & 74.2 & 69.6 \\

      WorldVLA~\cite{worldvla}
      & $\checkmark$ & -
      & 60.0 & 83.4 & 96.2 & 87.6 & 81.8
      & 0.1 & 27.9 & 41.6 & 43.7
      & 17.1 & 10.9 & 38.0 & 25.0 \\

      X-VLA~\cite{xvla}
      & $\checkmark$ & -
      & \textbf{97.6} & 97.8 & \textbf{98.6} & 98.2 & \textbf{98.1}
      & 23.4 & \textbf{89.7} & 75.7 & 88.2
      & \textbf{96.0} & 62.7 & 71.8 & 71.4 \\

      \midrule
      \multicolumn{16}{l}{\textit{WAM}} \\
      \midrule

      X-WAM~\cite{xwam}
      & $\checkmark$ & $100.0\%$
      & 95.6 & 95.4 & 98.6 & 96.8 & 96.6
      & \textbf{72.9} & 46.3 & 88.5 & 93.6
      & \textbf{88.3} & 62.9 & 80.8 & 74.8 \\

      Fast-WAM~\cite{fastwam}
      & $\times$ & $0.0\%$
      & 95.2 & 97.0 & \textbf{100.0} & 98.2 & 97.6
      & 18.8 & 45.7 & 70.1 & 83.2
      & 45.7 & 29.8 & 62.7 & 49.1 \\

      Fast-WAM-Joint~\cite{fastwam}
      & $\times$ & $100.0\%$
      & 96.8 & \textbf{98.2} & 99.4 & \textbf{99.6} & 98.5
      & 37.5 & 64.5 & 93.0 & 95.0
      & 55.9 & 47.3 & 79.6 & 66.3 \\

      Faster-WAM~\cite{fasterwam}
      & $\times$ & $100.0\%$
      & \textbf{98.2} & \textbf{98.2} & 99.8 & \textbf{99.6} & \textbf{99.0}
      & 53.8 & \textbf{71.6} & \textbf{94.7} & \textbf{96.3}
      & 61.3 & 63.6 & 79.1 & 73.6 \\

      \midrule
      \textbf{CompactWAM (Ours)}
      & $\times$ & $18.4\%$
      & 97.8 & 98.0 & 99.6 & 98.2 & 98.4
      & 66.8 & 51.2 & 92.1 & 96.1
      & 78.7 & \textbf{83.8} & \textbf{82.2} & \textbf{77.9} \\

      \bottomrule[0.9pt]
    \end{tabular}
  }
  
  \vspace{-0.4cm}
\end{table*}

\begin{table}[t]
  \caption{\textbf{Success rates (\%) on RoboTwin~2.0.} Policies trained on clean demonstrations are evaluated in clean and randomized scenes.}
  \label{tab:robotwin}
  \centering
  \small
  \setlength{\tabcolsep}{3pt}
  \renewcommand{\arraystretch}{0.90}

  \resizebox{\columnwidth}{!}{%
    \begin{tabular}{l|cc|ccc}
      \toprule[0.9pt]

      \multirow{2}{4.0cm}{Method}
      & \multirow{2}{*}{\shortstack{\textbf{Emb.}\\\textbf{PT.}}}
      & \multirow{2}{*}{\shortstack{Paradigm}}
      & \multicolumn{3}{c}{RoboTwin 2.0}\\

      \cmidrule(lr){4-6}

      & & &
      Clean & Random & Avg. \\

      \midrule

      $\pi_{0.5}$~\cite{pi05}
      & $\checkmark$ & VLA
      & \textbf{70.7} & \textbf{46.0} & \textbf{58.4} \\

      GalaxeaVLA~\cite{galaxeavla}
      & $\checkmark$ & VLA
      & 62.7 & 12.7 & 37.7\\

      X-VLA~\cite{xvla}
      & $\checkmark$ & VLA
      & 68.0 & 20.9 & 44.5\\
      
      Xiaomi-Robotics-0~\cite{xiaomi_robotics0}
      & $\checkmark$ & VLA
      & 62.9 & 18.2 & 40.5\\

      ABot-M0~\cite{abotm0}
      & $\checkmark$ & VLA
      & 57.4 & 30.4 & 43.9\\
      
      \midrule

      X-WAM~\cite{xwam}
      & $\checkmark$ &WAM
      & 70.0 & \textbf{25.8} & 47.9 \\
      Fast-WAM~\cite{fastwam}
      & $\times$ &WAM
      & 80.6 & 8.9 & 44.8 \\

      Fast-WAM-Joint~\cite{fastwam}
      & $\times$ &WAM
      & 82.9 & 17.8 & 50.4 \\
      
      Faster-WAM~\cite{fasterwam}
      & $\times$&WAM
      & \textbf{84.5} & 10.7 & 47.6 \\

      \midrule
      \textbf{CompactWAM (Ours)}
      & $\times$ &WAM
      & 82.8 & 23.2 & \textbf{53.0} \\

      \bottomrule[0.9pt]
    \end{tabular}
  }
  
  \vspace{-0.4cm}
\end{table}

\noindent\textbf{Benchmarks and protocols.}
We evaluate on LIBERO~\cite{libero}, LIBERO-Plus~\cite{libero_plus}, and RoboTwin~2.0~\cite{robotwin2}. For LIBERO, we use the Spatial, Object, Goal, and Long suites, each containing 10 tasks, and conduct about 2,000 evaluation trials across the 40 tasks. To assess generalization beyond the training distribution, we evaluate the LIBERO-trained policy on LIBERO-Plus without additional training. This benchmark tests seven perturbation categories: camera, robot, language, lighting, background, sensor noise, and layout. For RoboTwin~2.0, we train on 2,500 clean demonstrations across 50 bimanual tasks and evaluate clean and randomized scenes with 100 trials per task. We use the clean-to-randomized setting documented by the benchmark's public leaderboard~\cite{robotwin_leaderboard}.

\noindent\textbf{Implementation details.}
CompactWAM uses Wan2.2-TI2V-5B~\cite{wan,wan22} as its visual backbone. The sparse and dense pathways share the visual backbone, while their action experts are initialized from compatible backbone weights. The VAE and text encoder remain frozen. Each training sample contains a 32-step action chunk and nine video frames sampled with a temporal stride of four. We jointly optimize video and action flow-matching objectives together with the $\mathcal L_{\mathrm{FG}}$ and $\mathcal L_{\mathrm{DA}}$ losses of Dynamic Forcing. Training uses AdamW~\cite{adamw} with a learning rate of $10^{-4}$, weight decay of $10^{-2}$, a cosine learning-rate schedule, and bfloat16 precision, with a batch size of 16 on eight NVIDIA H20 GPUs. At inference, one visual pass encodes the current observation and compact dynamic tokens, caching keys and values for 10 action denoising steps.

\begin{table}[t]
  \caption{\textbf{Real-world task success rates (\%) under in-distribution and out-of-distribution conditions.} Avg. is the mean across the three tasks.}
  \label{tab:real_world}
  \centering
  \small
  \setlength{\tabcolsep}{3pt}
  \renewcommand{\arraystretch}{0.90}

  \resizebox{\columnwidth}{!}{%
    \begin{tabular}{l|cccc}
      \toprule[0.9pt]

      \multirow{2}{*}{Method}
      & \multicolumn{4}{c}{Real World} \\

      \cmidrule(lr){2-5}

      & Pick Fruits & Stack Blocks & Clean Desk & Avg. \\

      \midrule
      \multicolumn{5}{l}{\mbox{\textit{In-distribution}}} \\
      \midrule

      Fast-WAM-Joint~\cite{fastwam}
      & 88 & 70 & 56 & 71 \\

      Faster-WAM~\cite{fasterwam}
      & 80 & 74 & 72 & 75 \\

      \textbf{CompactWAM (Ours)}
      & \textbf{92} & \textbf{80} & \textbf{86} & \textbf{86} \\

      \midrule
      \multicolumn{5}{l}{\mbox{\textit{Out-of-distribution}}} \\
      \midrule

      Fast-WAM-Joint~\cite{fastwam}
      & 56 & 46 & 28 & 43 \\

      Faster-WAM~\cite{fasterwam}
      & 60 & 54 & 54 & 56 \\

      \textbf{CompactWAM (Ours)}
      & \textbf{88} & \textbf{66} & \textbf{68} & \textbf{74} \\

      \bottomrule[0.9pt]
    \end{tabular}%
  }
  
  \vspace{-0.6cm}
\end{table}

\begin{table*}[t]
  \caption{\textbf{Effect of Dynamic Forcing at a fixed compact dynamic token budget.} All variants retain dense-pathway objectives, in \#1, $\mathcal L_A^D$ trains the sparse action expert while blocking its gradients to the visual backbone.}
  \vspace{-0.2cm}
  \label{tab:abl}
  \centering
  \small
  \setlength{\tabcolsep}{3pt}
  \renewcommand{\arraystretch}{0.90}

  \resizebox{\textwidth}{!}{%
    \begin{tabular}{l|ccc|ccccc|cccccccc}
      \toprule[0.9pt]

      \multirow{2}{2.0cm}{Variants}
      & \multirow{2}{*}{\rule{0pt}{2.2ex}$\mathcal L_A^D$}
      & \multirow{2}{*}{$\mathcal L_{\mathrm{FG}}$}
      & \multirow{2}{*}{$\mathcal L_{\mathrm{DA}}$}
      & \multicolumn{5}{c|}{LIBERO}
      & \multicolumn{8}{c}{LIBERO-Plus} \\

      \cmidrule(lr){5-9}
      \cmidrule(l){10-17}

      & & & &
      Long & Goal & Object & Spatial & Avg.
      & Camera & Robot & Language & Light
      & Background & Noise & Layout & Avg. \\

      \midrule
      \# 1
      & $\times$ & $\times$& $\times$
      & 96.2 & 95.6 & 99.4 & 97.0 & 97.1
      & 49.0 & 37.0 & 77.5 & 90.8
      & 69.6 & 59.0 & 71.2 & 63.5 \\

      \# 2
      & $\checkmark$ & $\times$& $\times$
      & 94.2 & 94.2 & 99.8 & \textbf{98.4} & 96.7
      & 55.9 & 44.8 & 85.9 & 93.3
      & 65.4 & 73.1 & 76.8 & 70.0 \\

      \# 3
      & $\checkmark$ & $\times$ & $\checkmark$
      & \textbf{98.2} & 98.4 & 99.8 & 97.0 & 98.4
      & 63.5 & 45.4 & 90.5 & 96.3
      & 75.1 & 80.3 & 81.5 & 75.2 \\

      \# 4
      & $\checkmark$& $\checkmark$& $\times$
      & 97.8 & \textbf{98.8} & \textbf{100.0} & 97.4 & \textbf{98.5}
      & 62.2 & 49.4 & 90.6 & \textbf{98.8}
      & \textbf{81.8} & 77.5 & \textbf{82.4} & 76.3 \\

      \# 5
      & $\checkmark$ & $\checkmark$& $\checkmark$
      & 97.8 & 98.0 & 99.6 & 98.2 & 98.4
      & \textbf{66.8} & \textbf{51.2} & \textbf{92.1} & 96.1
      & 78.7 & \textbf{83.8} & 82.2 & \textbf{77.9} \\

      \bottomrule[0.9pt]
    \end{tabular}
  }
  
  \vspace{-0.4cm}
\end{table*}

\noindent\textbf{Metrics and comparisons.}
We report task success rates for evaluation. LIBERO averages the four task suites, whereas LIBERO-Plus aggregates successful trials over all evaluated perturbation trials, rather than taking an unweighted mean of the seven category rates. For RoboTwin~2.0, we report the mean task success rate in each setting and the average across the clean and randomized settings. Real-world results report per-task success rates and their rounded mean. We compare with the VLA and WAM methods cited in Tables~\ref{tab:Libero} and~\ref{tab:robotwin}, and the embodied-pretraining indicator distinguishes their training regimes. On RoboTwin~2.0, we reproduce Fast-WAM, Fast-WAM-Joint, and Faster-WAM using clean demonstrations and evaluate them under the same protocol as CompactWAM. We additionally compare against reported X-WAM results.

\subsection{Simulation Results}
\noindent\textbf{LIBERO and LIBERO-Plus.} Compact dynamic tokens preserve strong in-distribution control while improving robustness under distribution shift. As shown in Table~\ref{tab:Libero}, CompactWAM achieves 98.4\% on LIBERO and 77.9\% on LIBERO-Plus using only 18.8\% of the dense future token budget. It remains within 0.6 percentage points of Fast-WAM-Joint and Faster-WAM on LIBERO, yet exceeds these dense-future methods by 11.6 and 4.3 points on LIBERO-Plus, respectively. Against Fast-WAM, which removes future prediction at inference, retaining sparse dynamic context improves LIBERO-Plus success by 28.8 points. CompactWAM also surpasses the embodied-pretrained X-WAM by 1.8 points on LIBERO and 3.1 points on LIBERO-Plus. Together, these comparisons support compact dynamic tokens as an effective alternative to dense future generation. CompactWAM achieves the highest LIBERO-Plus average among the evaluated WAMs.

\begin{table}[t]
  \caption{\textbf{Inference-path comparison using the same benchmark-specific trained checkpoint.} Current-Only removes sparse dynamic context at inference, Sparse-Dynamic uses the learned sparse pathway, and Dense-Modeling evaluates the dense-future pathway and its action expert.}
  \label{tab:abl_state}
  \centering
  \small
  \setlength{\tabcolsep}{3pt}
  \renewcommand{\arraystretch}{0.90}

  \resizebox{\columnwidth}{!}{%
    \begin{tabular}{l|cc|cc}
      \toprule[0.9pt]

      Variants &  \shortstack{RoboTwin \\ Clean} & \shortstack{RoboTwin \\ Random} & LIBERO & LIBERO-Plus \\

      \midrule

      Current-Only & 3.1 & 0.4 & 22.1 & 5.7 \\
      Sparse-Dynamic & 82.8 & \textbf{23.2} & 98.4 & 77.9 \\
      \mbox{Dense-Modeling} & \textbf{84.6} & 21.3 & \textbf{98.9} & \textbf{79.4} \\
      \bottomrule[0.9pt]
    \end{tabular}
  }
  
  \vspace{-0.5cm}
\end{table}

\noindent\textbf{RoboTwin~2.0.} CompactWAM achieves the highest average success rate among the evaluated WAMs in Table~\ref{tab:robotwin}. It substantially improves performance in randomized scenes over the reproduced baselines while maintaining competitive performance in clean scenes.

\subsection{Real-World Results}
We evaluate \textit{Pick Fruits}, \textit{Stack Blocks}, and \textit{Clean Desk} under in-distribution and out-of-distribution conditions, using the same task data and matched evaluation conditions for CompactWAM, Fast-WAM-Joint, and Faster-WAM. Table~\ref{tab:real_world} shows that CompactWAM leads in all six task--condition pairs. Its rounded averages reach 86\% in distribution and 74\% out of distribution, exceeding Faster-WAM by 11 and 18 points, respectively. The larger advantage under distribution shift supports the utility of sparse dynamic context in the evaluated real-world tasks. Further details are provided in the supplementary material.

\begin{figure}[t]
  \centering
  \includegraphics[width=\columnwidth]{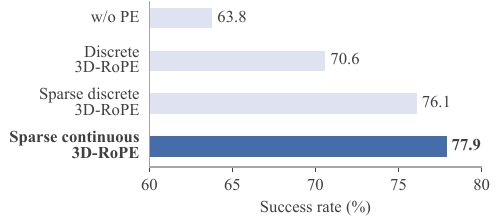}
  \vspace{-0.62cm}
  \caption{\textbf{Positional encoding for compact dynamic tokens on LIBERO-Plus.} We compare omitting positional encoding (w/o PE) against discrete, sparse discrete, and sparse continuous 3D-RoPE.}
  \vspace{-0.5cm}
  \label{fig:abl_pe}
\end{figure}

\subsection{Analysis of Sparse Dynamic Representations}

\noindent\textbf{Effect of Dynamic Forcing.}
Table~\ref{tab:abl} evaluates two sources of dynamic forcing within our sparse-dense framework. With the shared visual backbone supervised by dense video prediction, Variant \#1 trains the sparse-pathway action expert using $\mathcal{L}_A^D$ while detaching its gradients to the backbone. Enabling this gradient propagation (\#1 $\rightarrow$ \#2) provides direct action-driven forcing, improving LIBERO-Plus success from 63.5\% to 70.0\% while maintaining comparable LIBERO success at 96.7\% versus 97.1\%. This establishes the CompactWAM baseline with direct action-driven forcing and demonstrates the effectiveness of our framework for learning compact representations that support strong control performance. Building on this baseline, $\mathcal{L}_{\mathrm{FG}}$ and $\mathcal{L}_{\mathrm{DA}}$ provide additional dynamic forcing from the dense future pathway by transferring action-functional contributions and temporal relational changes, respectively. They individually raise LIBERO-Plus success to 76.3\% and 75.2\%, while their combination achieves 77.9\%, exceeding the baseline by 7.9 percentage points and the stronger individual objective by 1.6 points. These variants also improve LIBERO success to 98.4--98.5\%, supporting the complementary value of dense-future knowledge beyond direct action supervision.

\noindent\textbf{Role of predictive context.} The learned sparse context contributes substantially to action generation. Table~\ref{tab:abl_state} compares the deployed \textit{Sparse-Dynamic} pathway with \textit{Current-Only}, which removes compact dynamic tokens, and \textit{Dense-Modeling}, which uses dense future features and their corresponding action expert. Removing sparse context from the trained policy reduces success to 22.1\% on LIBERO and 5.7\% on LIBERO-Plus, with similarly severe drops on RoboTwin. Because this intervention involves no retraining, it demonstrates the policy's reliance on its learned dynamic context. In contrast, Sparse-Dynamic stays within 1.5 percentage points of Dense-Modeling on both LIBERO benchmarks and within 1.9 points in both RoboTwin settings. Sparse dynamic representations therefore retain much of the dense pathway's utility for control at deployment.

\begin{figure}[t]
  \centering
  \includegraphics[width=\columnwidth]{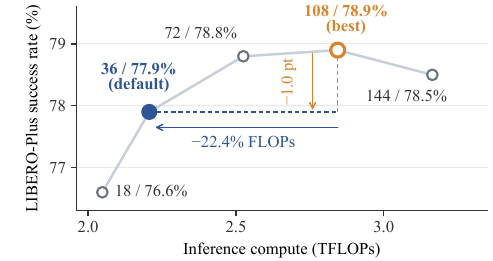}
  \vspace{-0.62cm}
  \caption{\textbf{Performance-compute trade-off on LIBERO-Plus.} Success rate vs. inference compute (TFLOPs) across compact dynamic token budgets. Blue marks the 36-token default, and orange marks the 108-token configuration with the highest success rate.}
  \vspace{-0.62cm}
  \label{fig:planning_budget}
\end{figure}

\noindent\textbf{Compact dynamic token budget.} Increasing representational capacity does not monotonically improve control. Figure~\ref{fig:planning_budget} compares budgets from 18 to 144 tokens on LIBERO-Plus. Even 18 tokens, just 9.2\% of the dense future budget, achieve 76.6\% success. Performance peaks at 108 tokens and dips slightly at 144 despite higher compute, indicating that additional capacity does not necessarily yield more useful predictive context. We use 36 tokens as a practical trade-off: 77.9\% success at 2.21 TFLOPs, within 1.0 point of the 108-token peak while requiring approximately 22\% less computation.

\noindent\textbf{Spatial organization of sparse tokens.} Preserving spatial structure is important even with a compact representation. In Figure~\ref{fig:abl_pe}, introducing discrete 3D-RoPE improves LIBERO-Plus success by 6.8 points over omitting positional encoding, and adapting the coordinates to the sparse token layout adds a further 5.5 points. Sparse continuous 3D-RoPE achieves 77.9\%, improving on its sparse discrete counterpart by 1.8 points. This comparison supports retaining continuous region-centroid coordinates when encoding compact dynamic tokens.

\subsection{Visualizing Sparse Dynamics}
\label{sec:dynamics_probe}
\begin{figure}[t]
  \centering
  \includegraphics[width=\columnwidth]{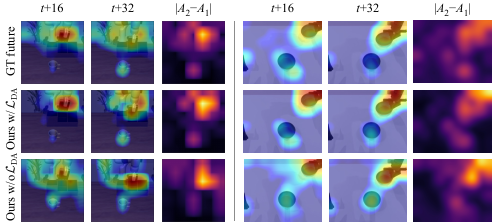}
  \vspace{-0.42cm}
  \caption{\textbf{Dynamics-focused attention on LIBERO (left) and RoboTwin (right).} Rows compare ground-truth future references and sparse representations learned with and without $\mathcal L_{\mathrm{DA}}$. Columns show attention maps $A_1$ at $t+16$ and $A_2$ at $t+32$, and their absolute difference $|A_2-A_1|$.}
  \vspace{-0.2cm}
  \label{fig:dynamics_probe}
\end{figure}

Figure~\ref{fig:dynamics_probe} compares dynamics-focused attention maps of sparse representations learned with and without $\mathcal L_{\mathrm{DA}}$ against ground-truth future references. In the illustrated examples, training with $\mathcal L_{\mathrm{DA}}$ yields temporal difference maps that more closely match the reference patterns in spatial location and extent. This qualitative correspondence supports its role in learning future temporal structure, complementing the success-rate ablations.

\begin{table}[!htbp]
  \caption{\textbf{Inference efficiency comparison.}
  Latency is reported in milliseconds as mean $\pm$ standard deviation.}
  \vspace{-0.2cm}
  \label{tab:comp_compute}
  \centering
  \footnotesize
  \setlength{\tabcolsep}{2pt}
  \renewcommand{\arraystretch}{0.85}

  \begin{tabular*}{\columnwidth}
    {@{\extracolsep{\fill}}l|ccc@{}}
    \toprule[0.9pt]
    Method & Latency & TFLOPs & Token Ratio \\
    \midrule
    Fast-WAM
      & $186.57 \pm 16.31$ & \textbf{3.59} & $\bm{\times 1.00}$ \\
    Fast-WAM-Joint
      & $537.54 \pm 14.10$ & 36.35 & $\times 3.00$ \\
    Faster-WAM
      & $172.79 \pm 11.30$ & 5.50 & $\times 3.00$ \\
    CompactWAM
      & $\bm{158.46 \pm 5.81}$ & 3.94 & $\times 1.53$ \\
    \bottomrule[0.9pt]
  \end{tabular*}
  \vspace{-0.4cm}
\end{table}

\subsection{Inference Efficiency}
\label{sec:inference_efficiency}
We benchmark inference, including language encoding, on an NVIDIA H20 GPU using the real-robot input format, 32-step action chunks, and 10 action denoising steps. Table~\ref{tab:comp_compute} shows that sparse dynamic context offers a favorable performance-efficiency trade-off. CompactWAM has the lowest mean latency at 158.46\,ms, with a $3.39\times$ speedup over Fast-WAM-Joint and a $1.09\times$ speedup over Faster-WAM. Relative to Faster-WAM, it cuts computation by 28.4\%, from 5.50 to 3.94 TFLOPs per action chunk, and reduces visual context from 360 to 184 tokens. Although it retains predictive context, CompactWAM is also $1.18\times$ faster than Fast-WAM in this setting. These efficiency gains complement the control and robustness results.

\section{Conclusion}
\label{sec:conclusion}

We presented CompactWAM for sparse dynamic representation learning in world action models. Its sparse-dense joint predictive architecture learns compact dynamic tokens within a video-supervised shared backbone, without a video reconstruction objective for the sparse pathway. Dynamic Forcing combines direct action supervision with privileged dense future guidance to promote action relevance and dynamic fidelity, with the latter matching action-functional contributions and temporal relational changes. At inference, compact dynamic tokens complement current-observation features for action prediction, eliminating dense future generation. Experiments in simulation and the real world demonstrate strong control performance and robustness under distribution shifts. With only 9.2\% of the dense future token budget, CompactWAM achieves success rates of 98.3\% on LIBERO and 76.6\% on LIBERO-Plus. These findings support a compact form of visual foresight: retaining rich scene context in current observations while learning sparse representations of anticipated changes relevant to robot control.
{
    \small
    \bibliographystyle{ieeenat_fullname}
    \bibliography{main}
}
\clearpage
\setcounter{page}{1}
\maketitlesupplementary
\setcounter{dbltopnumber}{2}

\providecommand{\supptodo}[1]{%
  \ifmmode\text{\textcolor{blue}{\textbf{[TBD:} #1\textbf{]}}}%
  \else\textcolor{blue}{\textbf{[TBD:} #1\textbf{]}}\fi}
\providecommand{\supptbd}{\textcolor{blue}{\textbf{TBD}}}
\providecommand{\suppfig}[2]{%
  \IfFileExists{#1}{\includegraphics[width=\linewidth]{#1}}{%
    \fbox{\parbox[c][29mm][c]{0.94\linewidth}{%
      \centering\small\textbf{Figure to add}\\[3pt]#2}}}}

\appendix
\setcounter{table}{0}
\setcounter{figure}{0}
\setcounter{equation}{0}
\renewcommand{\thetable}{S\arabic{table}}
\renewcommand{\thefigure}{S\arabic{figure}}
\renewcommand{\theequation}{S\arabic{equation}}
\ifdefined\theHtable\renewcommand{\theHtable}{supp.\arabic{table}}\fi
\ifdefined\theHfigure\renewcommand{\theHfigure}{supp.\arabic{figure}}\fi
\ifdefined\theHequation\renewcommand{\theHequation}{supp.\arabic{equation}}\fi

\noindent\textbf{Overview.}
This supplementary material presents real-world experimental details in
Section~\ref{sec:supp_real} and method and implementation details in
Section~\ref{sec:supp_method}. Section~\ref{sec:supp_analysis} examines
how compact dynamic tokens contribute to action generation and how their
representations vary across anticipated future times.
Section~\ref{sec:supp_more_findings} studies sensitivity to instruction
metadata and its relation to text--visual attention.

\section{Real-World Experimental Details}
\label{sec:supp_real}

\subsection{Platform and Data Collection}
\label{sec:supp_platform}

We conduct real-world experiments on an AGIBOT G2 dual-arm robot
with two-finger grippers. The policy receives RGB images from one
head-mounted and two wrist-mounted cameras, proprioceptive states,
and a language instruction. We collect 200 demonstrations per task
through PICO VR teleoperation, totaling 600 demonstrations across
Pick Fruits, Stack Blocks, and Clean Desk. CompactWAM,
Fast-WAM-Joint~\cite{fastwam}, and Faster-WAM~\cite{fasterwam}
are trained on the same demonstrations and evaluated under the same
task conditions.

All methods use ten action denoising steps and execute 16 actions
from each 32-step prediction at 30\,Hz before replanning. Model inputs
and training configurations are provided in
Section~\ref{sec:supp_method}.

\begin{figure*}[t]
  \centering
  \includegraphics[width=\textwidth]{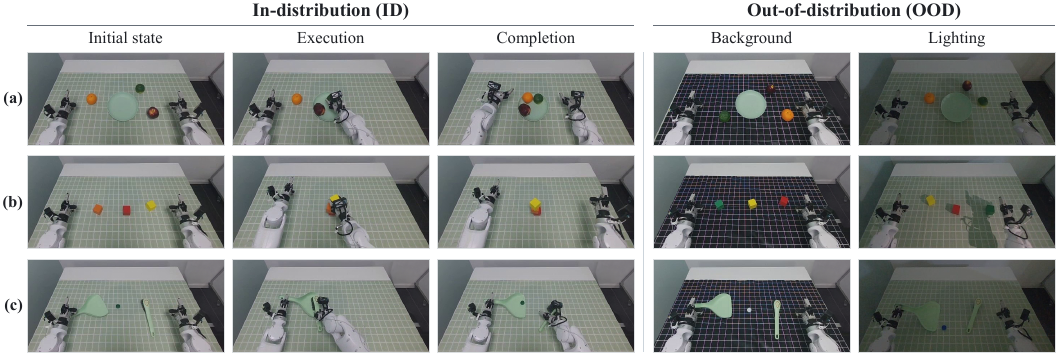}
  \caption{\textbf{Real-world tasks and evaluation conditions.}
  Each row shows an ID execution sequence and examples of background
  and lighting perturbations. The task instructions for the ID
  sequences are: \textbf{(a) Pick Fruits:} \textit{``Pick up
  the fruits and place them on the plate in this order: red, green,
  orange.''} \textbf{(b) Stack Blocks:} \textit{``Stack the blocks
  from bottom to top in this order: red, orange, yellow.''}
  \textbf{(c) Clean Desk:} \textit{``Clean the table.''}}
  \label{fig:supp_real_tasks}
\end{figure*}

\subsection{Tasks and Evaluation Settings}
\label{sec:supp_real_protocol}

Figure~\ref{fig:supp_real_tasks} illustrates the three tasks and
their evaluation conditions.

\noindent\textbf{Pick Fruits.}
The robot transfers all requested fruits to the designated plate
in the order specified by the instruction.

\noindent\textbf{Stack Blocks.}
The robot stacks the requested blocks in the specified color order.
Success requires the completed stack to remain standing after release.

\noindent\textbf{Clean Desk.}
The robot picks up the broom and dustpan, then coordinates both arms
to sweep the designated tabletop objects into the dustpan.

\noindent\textbf{Evaluation settings.}
In-distribution (ID) evaluation follows the demonstration setup, while
out-of-distribution (OOD) evaluation
either replaces the light-green tablecloth with a black one or
reduces ambient illumination to create a dark environment.
Each method is evaluated on 50 ID trials and 50 OOD trials per task,
with 25 trials for each OOD perturbation. We report full-task success
rates, combining both perturbations for the OOD score and averaging
across the three tasks for the overall results.

\subsection{Detailed OOD Results}
\label{sec:supp_real_ood}

\begin{figure}[!t]
  \centering
  \includegraphics[width=\columnwidth]{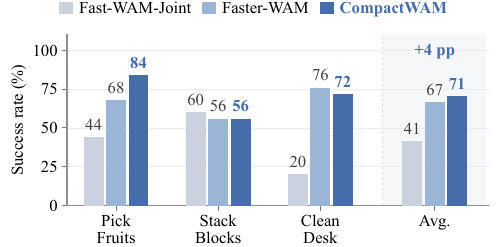}
  \caption{\textbf{Real-world success under background changes.}
  The tabletop is covered with a black tablecloth. Each task has
  25 trials. Avg. pools the 75 trials across tasks. The annotation
  marks the average gain over the stronger baseline.}
  \label{fig:supp_real_ood_background}
\end{figure}

\begin{figure}[!t]
  \centering
  \includegraphics[width=\columnwidth]{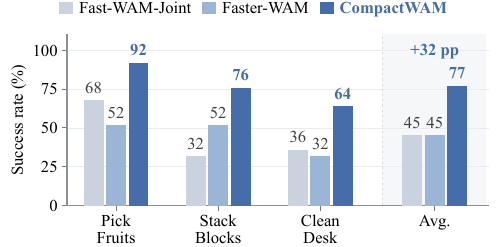}
  \caption{\textbf{Real-world success under low light.}
  Ambient illumination is reduced to create a dark environment.
  Each task has 25 trials. Avg. pools the 75 trials across tasks.
  The annotation marks the average gain over both baselines.}
  \label{fig:supp_real_ood_low_light}
\end{figure}

Figures~\ref{fig:supp_real_ood_background} and~\ref{fig:supp_real_ood_low_light}
break down the real-world OOD evaluation into black-tablecloth and low-light
conditions, with 25 trials per task and 75 trials per condition.
CompactWAM is more robust to both visual changes, with a particularly
large advantage under low light.

Under background changes, CompactWAM achieves the highest average success
rate of 71\%, exceeding Faster-WAM by 4 percentage points and Fast-WAM-Joint
by about 30 points. This advantage is driven by Pick Fruits, where it reaches
84\% success. Performance is closer on the other tasks: CompactWAM trails
Fast-WAM-Joint on Stack Blocks and Faster-WAM on Clean Desk by just one
successful trial each.

Under low light, CompactWAM leads on all three tasks, reaching 77\% on
average and outperforming both baselines by 32 percentage points.
Compared with the ID results in Table~3 of the main paper, its average
success drops by only 9 points, while both dense-future baselines lose
more than 25 points. This smaller degradation is consistent with retaining
interaction dynamics in a compact representation rather than modeling
dense appearance. Across all six task--perturbation pairs, CompactWAM
maintains at least 56\% success, avoiding the sharp drops observed for
the baselines on Clean Desk. It achieves the highest combined OOD
average of 74\%, with an 18-point lead over the stronger baseline.

\subsection{Qualitative Results}
\label{sec:supp_real_qual}

\begin{figure*}[t]
  \centering
  \includegraphics[width=\textwidth]{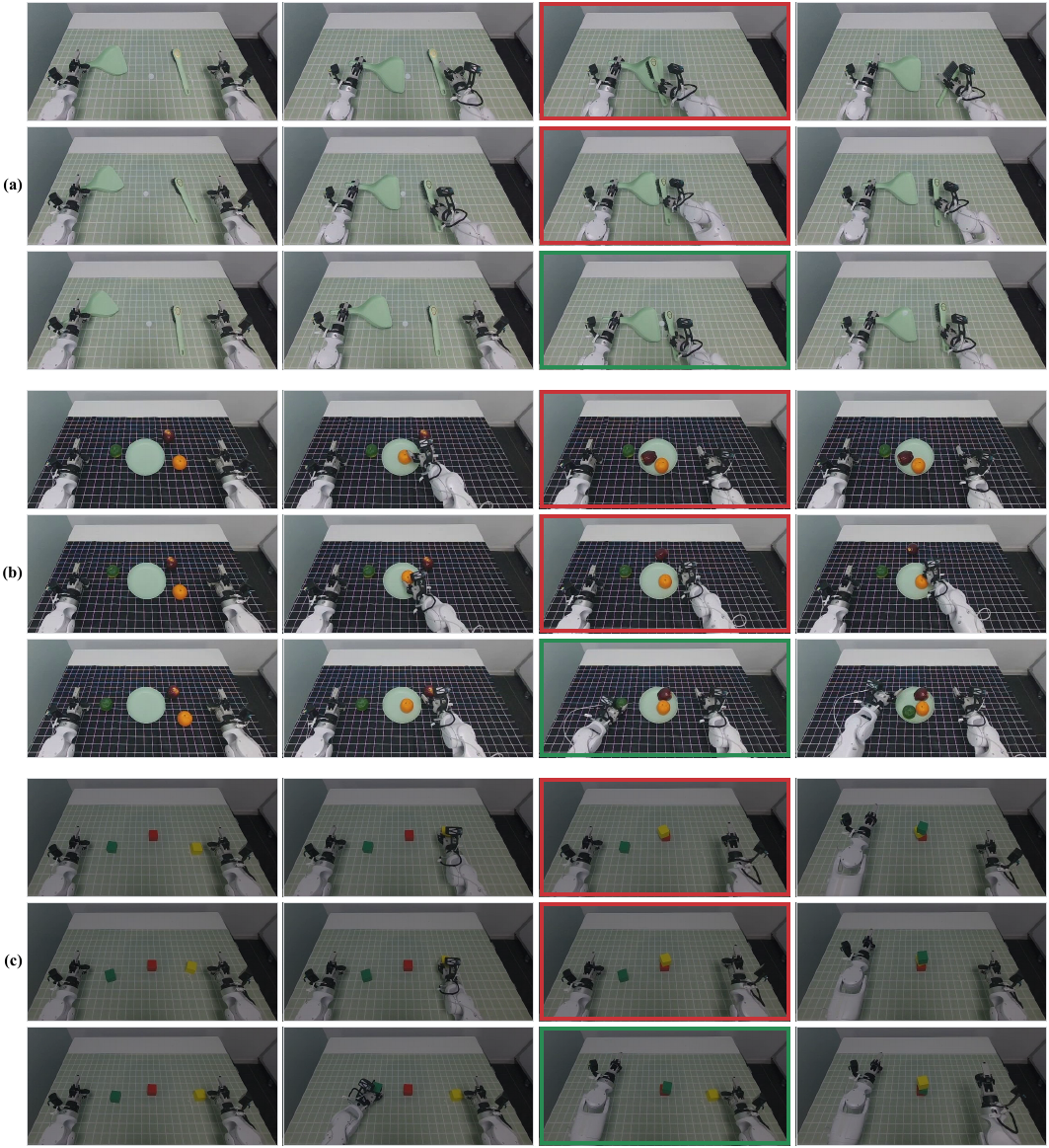}
  \caption{\textbf{Real-world execution comparisons under ID and OOD conditions.}
  \textbf{(a) Clean Desk, ID:} \textit{``Clean the table.''}
  \textbf{(b) Pick Fruits, background perturbation:} \textit{``Pick up
  the fruits and place them on the plate in this order: orange, red,
  green.''} \textbf{(c) Stack Blocks, lighting perturbation:}
  \textit{``Stack the blocks from bottom to top in this order: red,
  green, yellow.''} In each group, rows from top to bottom show
  Fast-WAM-Joint, Faster-WAM, and CompactWAM (Ours), with execution
  progressing from left to right. Red and green boxes highlight key
  moments in failed and successful executions, respectively.}
  \label{fig:supp_real_qual}
\end{figure*}

Figure~\ref{fig:supp_real_qual} compares task execution under ID,
background-shift, and low-light conditions, complementing the
quantitative results in Table~3 of the main paper. The examples
highlight differences in precise contact, continued task execution,
and instruction following. Together with the OOD breakdown in
Figures~\ref{fig:supp_real_ood_background} and~\ref{fig:supp_real_ood_low_light}, they illustrate how CompactWAM maintains
task completion and the requested action order under visual changes.

\noindent\textbf{Contact accuracy.}
In the ID Clean Desk example, both Fast-WAM-Joint and Faster-WAM
move the broom without making contact with the bottle cap, leaving
it outside the dustpan. CompactWAM brings the broom into contact
with the cap and sweeps it into the dustpan.

\noindent\textbf{Task progression under background changes.}
In Pick Fruits with a black tablecloth, Fast-WAM-Joint stalls after
picking and placing the red fruit and does not proceed to the green
fruit. Faster-WAM misaligns its grasp of the red fruit and fails to
pick it up. CompactWAM continues through the requested sequence
and places all three fruits on the plate.

\noindent\textbf{Instruction following under low light.}
In Stack Blocks, the instruction specifies red, green, and yellow
from bottom to top. Fast-WAM-Joint and Faster-WAM instead place the
yellow block below the green block, violating the requested order.
CompactWAM completes the stack in the specified order.

\section{Implementation Details}
\label{sec:supp_method}

We summarize the model configurations, training settings, and evaluation
protocols used in our experiments.

\subsection{Model and Token Configurations}
\label{sec:supp_architecture}

\noindent\textbf{Architecture.}
CompactWAM uses Wan2.2-TI2V-5B~\cite{wan,wan22}, with 30 visual blocks,
width 3,072, and 24 attention heads of dimension 128. Each action expert
has 30 blocks of width 1,024, using 24 heads at the eight visual
interaction blocks and eight heads elsewhere. All action blocks retain
text cross-attention, and proprioception is projected into an additional
conditioning token. The dense action expert is initialized from the
compact-pathway action expert, and both experts are jointly trained.
The VAE and text encoder are frozen.

\noindent\textbf{Input and token layouts.}
Nine frames sampled at offsets $0,4,\ldots,32$ are encoded into one
current and two future latent-time groups. The VAE downsamples images
spatially by a factor of 16, followed by patches of size $(1,2,2)$.
LIBERO combines two $224\times224$ views into a $224\times448$ image.
RoboTwin and real-world experiments use a $384\times320$ image comprising
a $256\times320$ head view above two $128\times160$ wrist views.
Table~\ref{tab:supp_tokens} summarizes the numbers of current-observation
tokens ($N_I$), dense future tokens ($N_F$), and compact dynamic tokens
($N_D$).

\begin{table}[t]
  \caption{\textbf{Visual token configurations.} Counts include all
  camera views and latent-time groups.}
  \label{tab:supp_tokens}
  \centering
  \small
  \begin{tabularx}{\columnwidth}{@{}Xccc@{}}
    \toprule
    Input configuration & $N_I$ & $N_F$ & $N_D$ \\
    \midrule
    LIBERO / LIBERO-Plus & 98 & 196 & 36 \\
    RoboTwin 2.0 & 120 & 240 & 64 \\
    Real-world manipulation & 120 & 240 & 64 \\
    \bottomrule
  \end{tabularx}
\end{table}

For each future latent-time group, compact dynamic tokens use regional
grids of $2\times3$ and $3\times4$ for the LIBERO agent and wrist views. RoboTwin and real-world experiments use $4\times5$, $2\times3$, and $2\times3$ grids for the head and two wrist views. The 18-token LIBERO variant uses $1\times3$ and $2\times3$ grids. The 36- and 18-token configurations retain 18.4\% ($36/196$) and 9.2\% ($18/196$) of the dense future token budget, respectively. All current-observation tokens are retained, giving $N_I + N_D$ visual tokens at inference.

\noindent\textbf{Sparse continuous 3D-RoPE.}
Each compact dynamic token inherits the spatial centroid of its source
region in the dense patch grid of the assembled multi-view image.
We retain the fractional coordinates in this original grid, preserving
both the spatial scale and the offsets between camera views. The same
spatial layout is used at future latent-time indices $t=1,2$.
These temporal, height, and width coordinates define sparse continuous
3D-RoPE, which is used consistently during training and inference.

The positional-encoding ablation in Figure~3 of the main paper compares
this configuration with three alternatives. Sparse discrete 3D-RoPE
rounds the region-centroid coordinates to the nearest integers.
Discrete 3D-RoPE uses integer coordinates on each view's compact grid,
starting at $(0,0)$ for each view. Both retain the future-time indices.
The w/o PE variant removes rotary positional encoding from the compact
dynamic tokens.

\subsection{Supervision Settings}
\label{sec:supp_forcing}

Both action experts receive the same noisy action sequence, timestep,
and flow-matching target. Their action losses update the shared visual
backbone through their respective conditioning keys and values.
The compact pathway receives action supervision and privileged guidance
from action functional guidance (FG) and dynamics-aware distribution
matching (DA), without a video reconstruction loss.

\noindent\textbf{Action functional guidance.}
FG uses blocks $\{0,4,8,12,16,20,24,28\}$, with zero-based indices.
Each pathway uses its own action query and attention denominator.
Both predictive contributions are mapped into the compact action expert's
residual space using its output projection and residual gate, then matched
with a Smooth-$\ell_1$ loss. The dense target, output projection, and residual
gate are detached.

\noindent\textbf{Dynamics-aware distribution matching.}
DA uses block-input features from $\{8,16,24,28\}$ and the future-time
pair $(1,2)$, computing relations separately for each camera.
Features undergo non-affine LayerNorm and $\ell_2$ normalization.
We use temperature $\tau=0.1$ and the region-size correction defined
in the main paper. The dense target distributions are detached for
KL-divergence matching.

\subsection{Training and Inference}
\label{sec:supp_training}

Tables~\ref{tab:supp_training} and~\ref{tab:supp_recipes} summarize the
common and benchmark-specific settings. Action and proprioception
dimensions are 7 and 8 for LIBERO, 14 and 14 for RoboTwin, and 16 and 16
for real-world experiments. LIBERO uses min--max normalization, while
RoboTwin and real-world experiments use z-score normalization with
training statistics. The video loss supervises future frames only.

\begin{table}[t]
  \caption{\textbf{Common training settings.} The same optimization
  setup is used for LIBERO, RoboTwin, and real-world experiments.}
  \label{tab:supp_training}
  \centering
  \small
  \begin{tabularx}{\columnwidth}{@{}lX@{}}
    \toprule
    Setting & Value \\
    \midrule
    Optimizer & AdamW~\cite{adamw} \\
    AdamW moments & $(0.9,0.95)$ \\
    Learning rate / weight decay & $10^{-4}$ / $10^{-2}$ \\
    Learning-rate schedule & Cosine, 5\% linear warm-up \\
    Gradient clipping & Global norm 1.0 \\
    Precision & bfloat16 \\
    Training hardware & 8 NVIDIA H20 GPUs \\
    Training seed & 42 \\
    \bottomrule
  \end{tabularx}
\end{table}

\begin{table}[t]
  \caption{\textbf{Benchmark-specific training settings.} FG and DA weights increase linearly over the listed warm-up steps. RoboTwin and real-world training use two gradient accumulation steps to reach an effective batch size of 256.}
  \label{tab:supp_recipes}
  \centering
  \small
  \setlength{\tabcolsep}{3pt}
  \begin{tabularx}{\columnwidth}{@{}Xccc@{}}
    \toprule
    Setting & LIBERO & RoboTwin & Real world \\
    \midrule
    Training epochs & 10 & 5 & 25 \\
    Effective batch size & 128 & 256 & 256 \\
    FG weight $\lambda$ & 0.5 & 0.1 & 0.1 \\
    DA weight $\beta$ & 0.5 & 0.5 & 0.5 \\
    FG / DA warm-up steps & 2000 & 1000 & 1000 \\
    \bottomrule
  \end{tabularx}
\end{table}

\noindent\textbf{Noise sampling and loss weights.}
Let $\psi(u)=5u/(1+4u)$ for $u\sim\mathcal U(0,1)$.
We set the visual noise level to $s=1$ with probability 0.5 and sample
$s=\psi(u)$ otherwise. The dense and compact pathways share $s$ but
use independent noise samples. Action noise levels are sampled
independently using $\psi$.
The video loss is weighted by 3. Table~\ref{tab:supp_recipes} lists
the FG and DA weights and their linear warm-up steps.

\noindent\textbf{Augmentation.}
Training uses random cropping, rotation, color and gamma jitter, and
Gaussian noise. Augmentation is applied with probability 0.65 on LIBERO
and 0.5 on RoboTwin and real-world data. LIBERO additionally uses
exposure jitter. No augmentation is applied during evaluation.

\noindent\textbf{Inference.}
A single visual pass encodes the current observation and compact dynamic
tokens at $s_{\mathrm{inf}}=0.999$. The resulting keys and values are
cached for ten Euler action denoising steps. Inference uses only the
compact pathway and its action expert.

\subsection{Evaluation Protocols}
\label{sec:supp_eval}

\noindent\textbf{Simulation.}
LIBERO~\cite{libero} uses 50 trials per task with benchmark-provided
initial states. Episode limits are 400 steps for Spatial, Object, and
Goal, and 700 for Long, following 30 settling steps. The policy executes
ten actions before replanning. LIBERO-Plus~\cite{libero_plus} evaluates
the LIBERO checkpoint without further training.
RoboTwin 2.0~\cite{robotwin2} uses 2,500 clean demonstrations across
50 tasks, with 100 evaluation trials per task in each clean and
randomized setting. The policy executes 24 actions before replanning.

LIBERO reports the average success rate over four suites. LIBERO-Plus
reports total successes divided by the total number of perturbation
trials. RoboTwin reports mean task success rates for the clean and
randomized settings and their average. Real-world tasks and evaluation
protocols are described in Section~\ref{sec:supp_real}.

\noindent\textbf{Inference efficiency.}
We measure inference on a single NVIDIA H20 GPU in bfloat16 with batch
size one, using the real-world configuration: $384\times320$ three-view
images, 16-dimensional proprioception, $32\times16$ action chunks, and
ten action denoising steps. CompactWAM uses 64 compact dynamic tokens.
Timing covers language encoding, current-image VAE encoding, visual
prefill, action sampling, and action output transfer to CPU.
After 128 warm-up requests, we collect 1,000 measurements per model.
We report latency as mean $\pm$ standard deviation.
Computation is measured per action chunk
over the same model components, counting one multiply--add as two FLOPs.

\section{Analysis of Compact Dynamic Representations}
\label{sec:supp_analysis}

We examine how compact dynamic tokens contribute to action prediction
and represent anticipated motion. Contribution and suppression analyses
measure their functional role, additional ablations compare supervision
targets and token budgets, and attention visualizations and future-frame
probes reveal the dynamics encoded in these representations.

\begin{table}[t]
  \caption{\textbf{Predictive contribution of compact dynamic tokens.}
  Predictive shares are reported as percentages for the compact and dense pathways.
  Supervision settings follow Table~4 of the main paper. In \#1,
  $\mathcal L_A^D$ trains the compact action expert with its gradients
  to the visual backbone stopped. Gain is the increase in compact
  predictive share over \#1, in percentage points.
  Differences are computed before rounding.}
  \label{tab:supp_future_share_single}
  \centering
  \small
  \setlength{\tabcolsep}{1.5pt}
  \renewcommand{\arraystretch}{1.15}
  \begin{tabular*}{\columnwidth}{@{\extracolsep{\fill}}ccccrrr@{}}
    \toprule
    \multirow{2}{*}{Variants}
      & \multicolumn{3}{c}{Supervision}
      & \multicolumn{2}{c}{Predictive share $\rho_Z$ (\%)}
      & \multirow{2}{*}{Gain} \\
    \cmidrule(lr){2-4}\cmidrule(lr){5-6}
      & $\mathcal L_A^D$ & $\mathcal L_{\mathrm{FG}}$ & $\mathcal L_{\mathrm{DA}}$ & Compact & Dense & \\
    \midrule
    \# 1 & $\times$ & $\times$ & $\times$ & \textbf{29.267} & 63.696 & --- \\
    \# 2 & $\checkmark$ & $\times$ & $\times$ & \textbf{36.481} & 63.286 & $\uparrow7.215$ \\
    \# 3 & $\checkmark$ & $\times$ & $\checkmark$ & \textbf{42.022} & 59.869 & $\uparrow12.756$ \\
    \# 4 & $\checkmark$ & $\checkmark$ & $\times$ & \textbf{43.361} & 60.242 & $\uparrow14.094$ \\
    \# 5 & $\checkmark$ & $\checkmark$ & $\checkmark$ & \textbf{44.382} & 59.865 & $\uparrow15.115$ \\
    \bottomrule
  \end{tabular*}
\end{table}

\begin{table*}[t]
  \caption{\textbf{Token contributions and action sensitivity.} The left group measures per-token contribution, and the right compares RMS action changes under Top-9 and matched Random-9 suppression. Per-token shares are reported as percentages. Diff.\ is compact minus dense, with share differences in percentage points. Variant definitions follow Table~\ref{tab:supp_future_share_single}. Differences are computed before rounding.}
  \label{tab:supp_supporting_branch_metrics}
  \centering
  \small
  \setlength{\tabcolsep}{1.5pt}
  \renewcommand{\arraystretch}{1.12}
  \begin{tabular*}{\textwidth}{@{\extracolsep{\fill}}c*{12}{r}@{}}
    \toprule
    \multirow{3}{*}{Variants}
      & \multicolumn{6}{c}{Token contribution}
      & \multicolumn{6}{c}{Action sensitivity} \\
    \cmidrule(lr){2-7}\cmidrule(lr){8-13}
      & \multicolumn{3}{c}{Per-token share $\bar s_Z$ (\%)}
      & \multicolumn{3}{c}{Contribution ratio $E_{\mathrm{pred}}$}
      & \multicolumn{3}{c}{Top-9 RMS}
      & \multicolumn{3}{c}{Random-9 RMS} \\
    \cmidrule(lr){2-4}\cmidrule(lr){5-7}
    \cmidrule(lr){8-10}\cmidrule(lr){11-13}
      & Compact & Dense & Diff.
      & Compact & Dense & Diff.
      & Compact & Dense & Diff.
      & Compact & Dense & Diff. \\
    \midrule
    \# 1 & 0.813 & 0.325 & $\uparrow0.488$ & 1.521 & 2.186 & $\downarrow0.664$ & 0.0446 & 0.1047 & $\downarrow0.0602$ & 0.0225 & 0.0148 & $\uparrow0.0077$ \\
    \addlinespace[2pt]
    \# 2 & 1.013 & 0.323 & $\uparrow0.690$ & 3.245 & 2.022 & $\uparrow1.223$ & 0.1485 & 0.1028 & $\uparrow0.0457$ & 0.0589 & 0.0145 & $\uparrow0.0444$ \\
    \addlinespace[2pt]
    \# 3 & 1.167 & 0.305 & $\uparrow0.862$ & 5.849 & 1.787 & $\uparrow4.062$ & 0.1931 & 0.1001 & $\uparrow0.0930$ & 0.0712 & 0.0129 & $\uparrow0.0583$ \\
    \addlinespace[2pt]
    \# 4 & 1.204 & 0.307 & $\uparrow0.897$ & 4.608 & 1.640 & $\uparrow2.968$ & 0.1747 & 0.0959 & $\uparrow0.0788$ & 0.0645 & 0.0126 & $\uparrow0.0519$ \\
    \addlinespace[2pt]
    \# 5 & 1.233 & 0.305 & $\uparrow0.927$ & 5.915 & 1.662 & $\uparrow4.252$ & 0.1753 & 0.0969 & $\uparrow0.0784$ & 0.0724 & 0.0133 & $\uparrow0.0592$ \\
    \bottomrule
  \end{tabular*}
\end{table*}

\begin{table*}[t]
  \caption{\textbf{Comparison of dense future supervision objectives.}
  Success rates (\%) on LIBERO and LIBERO-Plus compare direct feature
  alignment $\mathcal L_{\ell_2}$, cross-time matching $\mathcal L_{\mathrm{CT}}$,
  dynamics-aware distribution matching, and action functional guidance.}
  \label{tab:supp_guidance_objectives}
  \centering
  \small
  \setlength{\tabcolsep}{3pt}
  \renewcommand{\arraystretch}{0.90}

  \resizebox{\textwidth}{!}{%
    \begin{tabular}{l|ccccc|cccccccc}
      \toprule[0.9pt]

      \multirow{2}{2.0cm}{Variants}
      & \multicolumn{5}{c|}{LIBERO}
      & \multicolumn{8}{c}{LIBERO-Plus} \\

      \cmidrule(lr){2-6}
      \cmidrule(l){7-14}

      & 
      Long & Goal & Object & Spatial & Avg.
      & Camera & Robot & Language & Light
      & Background & Noise & Layout & Avg. \\

      \midrule
      
      $\mathcal L_{\mathrm{CT}}$
      & 97.2 & 98.0 & 99.6 & \textbf{99.0} & 98.4
      & 61.2 & 45.1 & 87.1 & 97.1
      & 76.5 & 79.4 & 81.3 & 74.4 \\

      $\mathcal L_{\ell_2}$
      & 96.8 & 97.8 & 99.8 & 98.2 & 98.1
      & 61.7 & \textbf{50.4} & \textbf{90.7} & 93.7
      & 69.8 & 79.9 & 78.7 & 74.4 \\
      \midrule
      $\mathcal L_{\mathrm{DA}}$
      & \textbf{98.2} & 98.4 & 99.8 & 97.0 & 98.4
      & \textbf{63.5} & 45.4 & 90.5 & 96.3
      & 75.1 & \textbf{80.3} & 81.5 & 75.2 \\

      $\mathcal L_{\mathrm{FG}}$
      & 97.8 & \textbf{98.8} & \textbf{100.0} & 97.4 & \textbf{98.5}
      & 62.2 & 49.4 & 90.6 & \textbf{98.8}
      & \textbf{81.8} & 77.5 & \textbf{82.4} & \textbf{76.3} \\

      \bottomrule[0.9pt]
    \end{tabular}
  }
  
\end{table*}

\begin{figure*}[t]
  \centering
  \includegraphics[width=\textwidth]{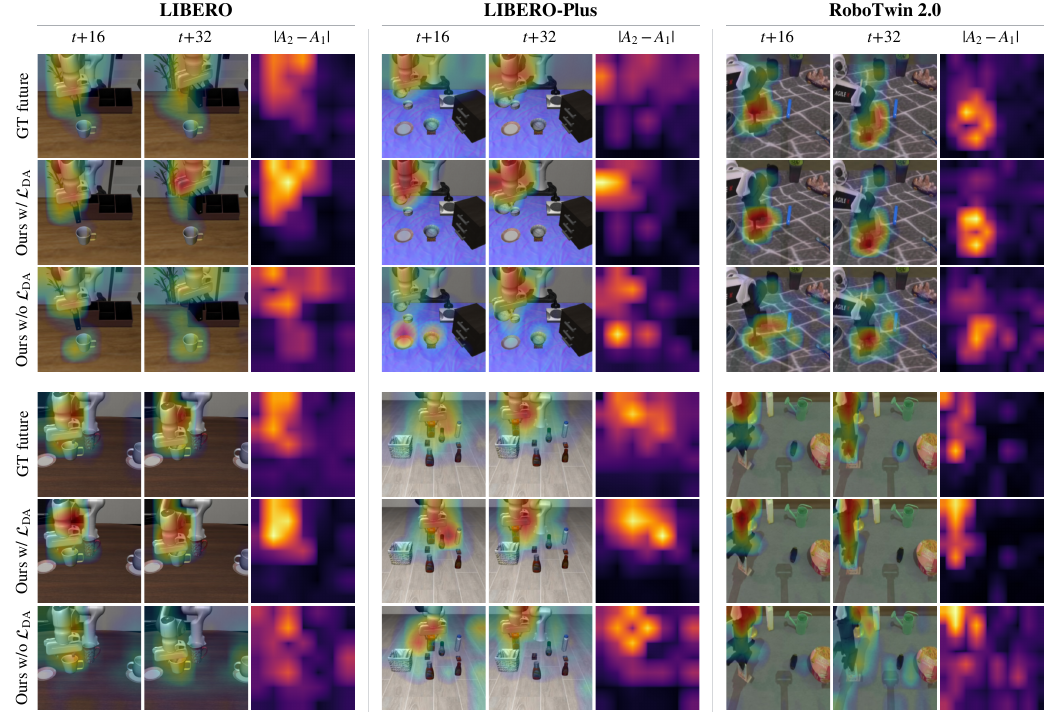}
  \vspace{-0.4cm}
  \caption{\textbf{Dynamics-focused attention across LIBERO, LIBERO-Plus, and RoboTwin~2.0.} Each benchmark column shows two examples. Rows compare ground-truth future references and compact dynamic representations learned with and without $\mathcal L_{\mathrm{DA}}$. Columns show attention maps $A_1$ at $t+16$ and $A_2$ at $t+32$, followed by their absolute difference $|A_2-A_1|$.}
  \label{fig:supp_compact_dynamics}
  \vspace{-0.4cm}
\end{figure*}

\subsection{Action Contributions and Sensitivity}
\label{sec:supp_analysis_protocol}
\label{sec:supp_functional}

\noindent\textbf{Analysis protocol.}
For Tables~\ref{tab:supp_future_share_single}
and~\ref{tab:supp_supporting_branch_metrics}, we use recorded demonstration
windows from all 40 LIBERO tasks: four episodes per task for contribution
analysis and two for token suppression, giving 160 and 80 windows,
respectively. The five variants use matched observations, instructions,
and three inference seeds. Contributions are measured using each
pathway's own action expert across all interaction blocks and ten action
denoising steps. Ratios are formed per block and step, then averaged
over seeds and windows with equal task weights.

\noindent\textbf{Predictive-token contributions.}
Let $Z\in\{D,F\}$ denote compact dynamic or dense future tokens and $I$
denote current-observation tokens. At each interaction block and
denoising step, we measure a token's contribution to its pathway's
gated action-attention update:
\begin{equation}
  \begin{aligned}
    \bm r_{qj}&=\operatorname{Concat}_{h}\!\left[
        \alpha_{qj}^{h}\bm v_j^{h}\right]
        \bm W_O^\top\odot\bm\gamma_q,\\
    c_j&=\frac{1}{|\mathcal Q|}
      \sum_{q\in\mathcal Q}\left\|\bm r_{qj}\right\|_2.
  \end{aligned}
  \label{eq:supp_contribution}
\end{equation}
Here $h$ indexes attention heads and $\mathcal Q$ is the set of action
queries. The attention probabilities
$\alpha$, values $\bm v$, output projection $\bm W_O$, and residual gate
$\bm\gamma$ belong to the corresponding action expert, retaining its
softmax over current-observation, predictive, and action keys.

Writing $C_Z=\sum_{j\in Z}c_j$ and $C_I=\sum_{j\in I}c_j$, we define
predictive share and mean share per predictive token as
\begin{equation}
  \rho_Z=\frac{C_Z}{C_Z+C_I+\varepsilon},
  \qquad \bar s_Z=\frac{\rho_Z}{N_Z},
  \label{eq:supp_share}
\end{equation}
where $N_Z=|Z|$ and $\varepsilon$ stabilizes the denominator. We also
measure the per-token contribution ratio relative to current-observation tokens:
\begin{equation}
  E_{\mathrm{pred}}=\frac{C_Z/N_Z}{C_I/N_I}.
  \label{eq:supp_enrichment}
\end{equation}
Both shares are reported as percentages, and $E_{\mathrm{pred}}$ is a
dimensionless ratio. Values above one indicate greater contribution per predictive token
than per current-observation token. Here $N_I=98$, $N_D=36$, and
$N_F=196$, so compact dynamic tokens occupy 26.9\% of the compact
pathway's visual context.

\noindent\textbf{Sensitivity to token suppression.}
\label{sec:supp_interventions}
We rank predictive tokens by their contribution share within $Z$,
averaged across blocks and denoising steps, and select the nine
highest-scoring tokens from the unperturbed prediction. Their values
are set to zero at every interaction block, retaining the original
keys and action noise. Random-9 averages ten random selections with
the same per-view and future-time token counts as Top-9. For either
selection $\mathcal M$, we measure the root-mean-square (RMS) change in the generated
action chunk:
\begin{equation}
  \Delta_A(\mathcal M)=
  \frac{\|\widetilde{\bm A}^{\mathcal M}
             -\widetilde{\bm A}\|_F}{\sqrt{H d_a}},
  \label{eq:supp_action_change}
\end{equation}
where $\widetilde{\bm A}$ and $\widetilde{\bm A}^{\mathcal M}$ are the
original and intervened predictions in training-normalized action
coordinates, with horizon $H=32$ and action dimension $d_a=7$.

\noindent\textbf{Effects of Dynamic Forcing.}
\label{sec:supp_quantitative}
All five variants retain dense video and action supervision and train
the compact action expert. Variant~\#1 stops compact action-loss gradients
at the visual backbone, while Variant~\#2 enables them. Variants~\#3 and~\#4
add DA and FG, respectively, and Variant~\#5 uses both.

Compact dynamic tokens contribute disproportionately to action
prediction. In the full model, they account for 44.4\% of visual
readout contribution with only 26.9\% of the visual tokens
(Table~\ref{tab:supp_future_share_single}). Enabling action gradients
increases both contribution share and suppression sensitivity, and privileged dense guidance further strengthens this functional use of
the compact context. The combined configuration has the highest
contribution share, while DA alone produces the greatest Top-9 sensitivity.

Top-9 suppression changes action predictions more than matched Random-9
suppression across all variants and both pathways
(Table~\ref{tab:supp_supporting_branch_metrics}). In the full compact
pathway, the RMS change is 0.1753 under Top-9 and 0.0724 under Random-9.
This difference shows that the contribution ranking identifies tokens
with greater influence on action prediction, beyond the effect of
suppressing an equal number of predictive tokens.

\begin{figure*}[t]
    \centering
    \includegraphics[width=\textwidth]{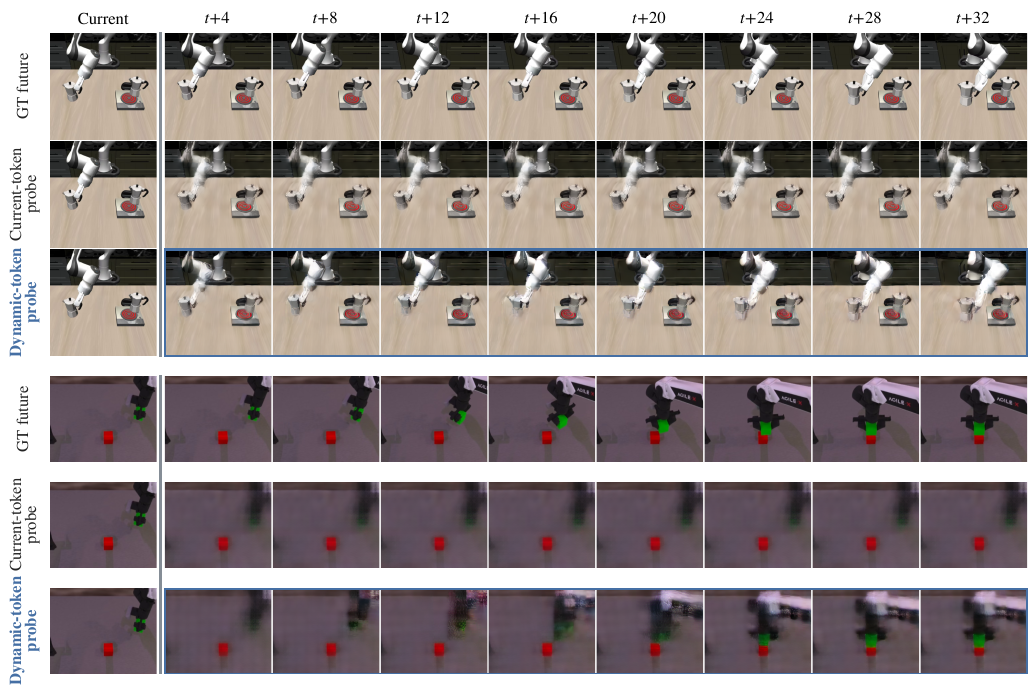}
    \caption{
    \textbf{Probing future dynamics encoded in compact dynamic tokens.}
    The upper and lower groups show examples from LIBERO and
    RoboTwin~2.0, respectively. The Current column repeats the same original
    input observation in all three rows. From $t+4$ to $t+32$, the rows show
    ground-truth future frames and predictions from probes conditioned on
    current-observation tokens (Current-token probe) and compact dynamic tokens
    (Dynamic-token probe), respectively. Both probes use the same network
    architecture.
    }
    \label{fig:supp_dynamic_probe}
\end{figure*}

\subsection{Additional Ablation Results}
\label{sec:supp_guidance_design}

\noindent\textbf{Dense future supervision objectives.}
Table~\ref{tab:supp_guidance_objectives} compares FG and DA with two
alternatives. Direct feature alignment $\mathcal L_{\ell_2}$ matches
compact features $H_t^D$ to regionally pooled dense features $\bm B H_t^F$, and cross-time matching $\mathcal L_{\mathrm{CT}}$ aligns affinities without
a static reference. FG and DA both improve LIBERO-Plus performance
over these alternatives while maintaining similar LIBERO success.
FG improves success by up to 1.9 percentage points, supporting
action-functional contributions and temporal relational changes as
targets for compact representation learning.

\noindent\textbf{Compact dynamic token budget.}
\label{sec:supp_token_capacity}
Table~\ref{tab:abl_sparsenum} extends the token-budget comparison, retaining
all current-observation tokens. LIBERO success varies little across the
evaluated budgets. On LIBERO-Plus, the 108-token configuration achieves
78.9\%, but increasing capacity to 144 tokens provides no further gain.
The 36-token default remains within one percentage point of the highest
success rate using one third as many compact dynamic tokens, balancing
predictive capacity and computation.

\begin{table}[!b]
  \caption{\textbf{Capacity of compact dynamic representations.} Average success rates (\%) for different compact dynamic token budgets.}
  \label{tab:abl_sparsenum}
  \centering
  \small
  \renewcommand{\arraystretch}{1.10}
  \begin{tabular*}{\columnwidth}{@{\extracolsep{\fill}}ccc@{}}
    \toprule
    Compact dynamic tokens & LIBERO Avg. & LIBERO-Plus Avg. \\
    \midrule
    18 & 98.3 & 76.6 \\
    36 & 98.4 & 77.9 \\
    72 & \textbf{98.6} & 78.8 \\
    108 & 98.3 & \textbf{78.9} \\
    144 & 98.5 & 78.5 \\
    \bottomrule
  \end{tabular*}
  \vspace{-0.2cm}
\end{table}

\begin{table*}[!t]
  \caption{\textbf{Effect of instruction-suffix correction across perturbations.} LIBERO-Plus success rates (\%) with original and corrected instructions. $\Delta$ reports the change in Avg. (corrected minus original), in percentage points.
  Fast-WAM, Fast-WAM-Joint, and Faster-WAM are reproduced with the same training
  augmentations as CompactWAM (Section~\ref{sec:supp_training}), so their original
  scores differ from the reported numbers in Table~1 of the main paper.
  Differences are computed before rounding.}
  \vspace{-0.2cm}
  \label{tab:supp_prompt_correction}
  \centering
  \small
  \setlength{\tabcolsep}{2pt}
  \begin{tabularx}{\textwidth}{@{}ll*{9}{>{\centering\arraybackslash}X}@{}}
    \toprule
    Method & Instruction & \makebox[\linewidth][c]{Camera} & \makebox[\linewidth][c]{Robot} & \makebox[\linewidth][c]{Language} & \makebox[\linewidth][c]{Light} & \makebox[\linewidth][c]{Background} & \makebox[\linewidth][c]{Noise} & \makebox[\linewidth][c]{Layout} & \makebox[\linewidth][c]{Avg.} & \makebox[\linewidth][c]{$\Delta$ (pp)} \\
    \midrule
    \multirow{2}{*}{Fast-WAM~\cite{fastwam}} & Original & 53.7 & 37.0 & 75.3 & 87.0 & 69.8 & 65.0 & 69.3 & 64.1 & \multirow{2}{*}{$+16.0$} \\
      & Corrected & \textbf{79.2} & 60.5 & 75.3 & 97.4 & 80.0 & 91.2 & 81.7 & 80.2 & \\
    \addlinespace[2pt]
    \multirow{2}{*}{Fast-WAM-Joint~\cite{fastwam}} & Original & 42.6 & 44.3 & 84.1 & 87.2 & 56.7 & 60.5 & 73.6 & 63.4 & \multirow{2}{*}{$+8.8$} \\
     & Corrected & 55.9 & 59.8 & 84.1 & 88.2 & 62.2 & 77.0 & 79.5 & 72.1 & \\
    \addlinespace[2pt]
    \multirow{2}{*}{Faster-WAM~\cite{fasterwam}} & Original & 59.8 & \textbf{58.6} & 89.5 & 96.3 & 67.3 & 73.2 & 79.5 & 74.3 & \multirow{2}{*}{$+6.8$} \\
     & Corrected & 69.7 & \textbf{73.0} & 89.5 & 96.6 & 71.8 & 86.1 & 82.0 & 81.0 & \\
    \addlinespace[2pt]
    \multirow{2}{*}{ImageWAM (FLUX.2 4B)~\cite{imagewam}} & Original & \textbf{80.4} & 47.9 & 91.4 & \textbf{96.8} & \textbf{86.7} & \textbf{94.3} & 78.5 & \textbf{81.5} & \multirow{2}{*}{$+0.5$} \\
     & Corrected & \textbf{79.2} & 49.9 & 91.4 & 96.0 & \textbf{89.8} & \textbf{94.8} & 78.8 & 82.0 & \\
    \addlinespace[2pt]
    \multirow{2}{*}{CompactWAM} & Original & 66.8 & 51.2 & \textbf{92.1} & 96.1 & 78.7 & 83.8 & \textbf{82.2} & 77.9 & \multirow{2}{*}{$+6.1$} \\
     & Corrected & 76.6 & 64.5 & \textbf{92.1} & \textbf{97.5} & 85.7 & 91.9 & \textbf{84.1} & \textbf{84.0} & \\
    \bottomrule
  \end{tabularx}
  \vspace{-0.2cm}
\end{table*}

\begin{table*}[t]
  \caption{\textbf{Examples of instruction-suffix correction.} Monospaced text marks scene metadata removed from the original instructions.}
  \label{tab:supp_prompt_examples}
  \centering
  \small
  \setlength{\tabcolsep}{5pt}
  \renewcommand{\arraystretch}{1.12}
  \begin{tabularx}{\textwidth}{@{}l*{2}{>{\raggedright\arraybackslash}X}@{}}
    \toprule
    Perturbation & Original instruction & Corrected instruction \\
    \midrule
    Light & pick up the black bowl between the plate and the ramekin and place it on the plate \texttt{light 31} & pick up the black bowl between the plate and the ramekin and place it on the plate \\
    \addlinespace[3pt]
    Camera & pick up the alphabet soup and place it in the basket \texttt{view 2 15 100 0 0 initstate 0} & pick up the alphabet soup and place it in the basket \\
    \addlinespace[3pt]
    Layout & put the wine bottle on top of the cabinet \texttt{moved level3 sample2} & put the wine bottle on top of the cabinet \\
    \bottomrule
  \end{tabularx}
\end{table*}

\subsection{Dynamics-Focused Attention}
\label{sec:supp_temporal_attention}

Figure~\ref{fig:supp_compact_dynamics} extends the main paper's
visualizations to LIBERO, LIBERO-Plus, and RoboTwin~2.0. The maps are
softmax-normalized query--key affinities between compact dynamic tokens
and features extracted from the corresponding recorded future frames.
The ground-truth future reference uses region-pooled queries extracted by the visual backbone from encoded ground-truth future frames, with keys from the same frames.

With $\mathcal L_{\mathrm{DA}}$, temporal difference maps more closely
follow the ground-truth future reference around the operating arm and manipulated
objects. Without $\mathcal L_{\mathrm{DA}}$, responses are more dispersed or extend into
static scene regions. These examples illustrate how dynamics-aware
distribution matching directs compact representations toward temporal
changes associated with interaction.

\subsection{Probing Compact Dynamic Tokens}
\label{sec:supp_dynamic_probe}

We compare future-frame probes conditioned on current-observation
tokens and compact dynamic tokens. Both use the same architecture
and differ only in their conditioning features from the frozen policy.
A two-layer Transformer decoder with width 512 and eight attention
heads reads these features through cross-attention from queries spanning
the future latent patch grid. The predicted latents are decoded into
eight RGB frames. The probes are fitted on training demonstrations using an MSE loss on future VAE latents, while the policy and VAE remain frozen. We visualize held-out episodes.

In Figure~\ref{fig:supp_dynamic_probe}, the dynamic-token probe recovers
the gripper-held moka pot moving toward the stove in LIBERO and the
green block descending onto the red block in RoboTwin~2.0. Their motion
direction and progression follow the ground-truth sequences, while the
current-token probe shows little temporal variation. These reconstructions
complement the attention maps by showing that anticipated, action-relevant
dynamics can be read out from frozen compact dynamic tokens.

\section{More Findings}
\label{sec:supp_more_findings}

\subsection{Sensitivity to Instruction Metadata}
\label{sec:supp_prompt_metadata}

In the evaluated LIBERO-Plus~\cite{libero_plus} release, non-language
instructions retain scene metadata from task filenames.
Table~\ref{tab:supp_prompt_examples} illustrates these suffixes and
their removal. The extra text couples evaluations of visual and physical
robustness with sensitivity to irrelevant instruction content.

For a fair comparison, we reproduce Fast-WAM, Fast-WAM-Joint, and
Faster-WAM using the same training augmentations as CompactWAM,
as described in Section~\ref{sec:supp_training}. We evaluate
ImageWAM~\cite{imagewam} using its official checkpoint, which was
also trained with data augmentation.

Each model uses the same checkpoint and task set for both instruction
versions, with one episode per variant. Correction affects 8,493 of
10,030 variants, excluding the 1,537 variants in the Language category.


\noindent\textbf{Suffix sensitivity varies across models and perturbations.}\par
\nopagebreak[4]\noindent
Table~\ref{tab:supp_prompt_correction} shows gains of
6.1--16.0 percentage points for video-based models, with Fast-WAM
benefiting most, compared with 0.5 points for ImageWAM~\cite{imagewam}.
CompactWAM achieves the highest corrected average of 84.0\%, exceeding ImageWAM by 2 percentage points.

\subsection{Joint Attention and Instruction Robustness}
\label{sec:supp_text_integration}

ImageWAM uses joint attention between text and visual
tokens, whereas our Wan-based backbone incorporates language through separate cross-attention. Joint attention normalizes over textual and visual keys together, whereas cross-attention normalizes over text keys alone. We hypothesize that this difference contributes to ImageWAM's lower sensitivity to the appended metadata.



\end{document}